\documentclass{article}
\usepackage{fullpage}
\usepackage[svgnames,dvipsnames,color,table,xcdraw]{xcolor}
\usepackage{hyperref}
\hypersetup{colorlinks,linkcolor={blue},citecolor={blue},urlcolor={RoyalBlue}}
\usepackage{subcaption}
\usepackage{authblk}

\usepackage{comment}

\usepackage{graphicx}
\usepackage{amsmath}
\usepackage{amssymb}
\usepackage{booktabs}
\usepackage{multirow}
\usepackage[round]{natbib}

\newcommand\sI{\ensuremath{\mathcal{I}}}

\newcommand\sT{\ensuremath{\mathcal{T}}}

\newcommand\sY{\ensuremath{\mathcal{Y}}}

\newcommand\R{\ensuremath{\mathbb{R}}} 

\DeclareMathOperator*{\argmax}{arg\,max} 
\usepackage[algo2e]{algorithm2e}

\usepackage{float}
\usepackage{caption}
\usepackage{algorithm}
\usepackage{algcompatible}

\usepackage[capitalize]{cleveref}
\crefname{section}{Sec.}{Secs.}
\Crefname{section}{Section}{Sections}
\Crefname{table}{Table}{Tables}
\crefname{table}{Tab.}{Tabs.}

\begin{document}
\looseness=-1

\linepenalty=1000

\title{Finetune like you pretrain: Improved finetuning of zero-shot vision models}

\author[1]{Sachin Goyal}
\author[2]{Ananya Kumar}
\author[1]{Sankalp Garg}
\author[1,3]{Zico Kolter}
\author[1]{Aditi Raghunathan}

\affil[1]{Carnegie Mellon University}
\affil[2]{Stanford University}
\affil[3]{Bosch Center for AI}

\maketitle

\begin{abstract}
   Finetuning image-text models such as CLIP achieves state-of-the-art accuracies on a variety of benchmarks. However, recent works ~\citep{wiseft, lpft} have shown that even subtle differences in the finetuning process can lead to surprisingly large differences in the final performance, both for in-distribution (ID) and out-of-distribution (OOD) data. In this work, we show that a natural and simple approach of mimicking contrastive pretraining consistently outperforms alternative finetuning approaches. Specifically, we cast downstream class labels as text prompts and continue optimizing the contrastive loss between image embeddings and class-descriptive prompt embeddings (contrastive finetuning). 
   
   Our method consistently outperforms baselines across 7 distribution shift, 6 transfer learning, and 3 few-shot learning benchmarks. On WILDS-iWILDCam, our proposed approach FLYP outperforms the top of the leaderboard by $2.3\%$ ID and $2.7\%$ OOD, giving the highest reported accuracy. Averaged across 7 OOD datasets (2 WILDS and 5 ImageNet associated shifts), FLYP gives gains of $4.2\%$ OOD over standard finetuning and outperforms the current state of the art (LP-FT) by more than $1\%$ both ID and OOD.  Similarly, on 3 few-shot learning benchmarks, our approach gives gains up to $4.6\%$ over standard finetuning and $4.4\%$ over the state of the art. In total, these benchmarks establish contrastive finetuning as a simple, intuitive, and state-of-the-art approach for supervised finetuning of image-text models like CLIP. Code is available at
\url{https://github.com/locuslab/FLYP}.

\end{abstract}

\newtheorem{remark}{Remark}
\newcommand{\conjf}[1]{\ensuremath{{#1}^\star}}
\newcommand{\f}{f_\theta}
\newcommand{\g}{g_\phi}
\newcommand{\h}{h}
\newcommand{\iminp}{x}
\newcommand{\Errid}{Err_\text{id}}
\newcommand{\Errood}{Err_\text{ood}}
\newcommand{\tinp}{t}
\newcommand{\simi}{\text{sim}}
\newcommand{\Lcont}{\mathcal{L}_\text{cont}}
\newcommand{\clip}{CLIP }
\newcommand{\zeroshot}{zero-shot }
\newcommand{\dembed}{{d_\text{e}}}
\newcommand{\hzs}{h_\text{zs}}
\newcommand{\temp}{\mathcal{T}}
\newcommand{\numtemp}{N_\text{template}}
\newcommand{\stemplate}{\{\mathcal{T}_j\}_{j=1}^t}
\newcommand{\iwildcam}{iWILDCam }
\newcommand{\fmow}{FMOW }

\newcommand{\dtrain}{\mathcal{D}_\text{train}}
\newcommand{\dtraintext}{\dtrain^\text{im-text}}
\newcommand{\numtrain}{N_\text{train}}
\newcommand{\Pid}{P_\text{id}}
\newcommand{\Pood}{P_\text{ood}}

\newcommand{\ftext}{g}
\newcommand{\fimg}{f}
\newcommand{\nftext}{\bar{g}}
\newcommand{\nfimg}{\bar{f}}
\newcommand{\hclass}{h_\text{class}}
\newcommand{\zshead}{h_\text{zs}}

\newcommand{\thetatext}{\theta_\text{text}}
\newcommand{\thetaimg}{\theta_\text{img}}
\newcommand{\lpretrain}{\mathcal{L}_\text{pre}}
\newcommand{\ours}{FLYP}
\newcommand{\method}{FLYP }

\section{Introduction}\label{sec:introduction}
Recent large-scale models pretrained jointly on image and text data, such as CLIP \citep{clip} or ALIGN \citep{align}, have demonstrated exceptional performance on many zero-shot classification tasks. These models are pretrained via a contrastive loss that finds a joint embedding over the paired image and text data. Then, for a new classification problem, one simply specifies a prompt (or, more commonly, a set of prompts) that describes the different classes. Finally, one can embed candidate images and predict the class with the highest similarity to the text embedding of the corresponding prompt. Such ``zero-shot'' classifiers involve no additional training at all and achieve reasonable performance on downstream tasks and impressive robustness to many common forms of distribution shift. However, in many cases, it is desirable to further improve performance via supervised finetuning: further training and updates to the pretrained parameters on a (possibly small) number of labeled images from the classification task.

In practice, however, several studies have found that standard finetuning procedures, while improving in-distribution performance, come at a cost to robustness to distribution shifts. Subtle changes to the finetuning process could mitigate this decrease in robustness. For example, ~\citet{lpft} demonstrated the role of initialization of the final linear head and proposed a two-stage process of linear probing, \emph{then} finetuning that achieves better than either method in isolation. As another example,~\citet{wiseft} showed that ensembling the weights of the finetuned and zero-shot classifier can improve robustness while simultaneously maintaining high in-distribution performance. Understanding the role of these subtle changes is challenging, and there is no simple recipe for what is the ``correct'' modification.

\begin{figure*}
    \centering
    \includegraphics[scale=0.43]{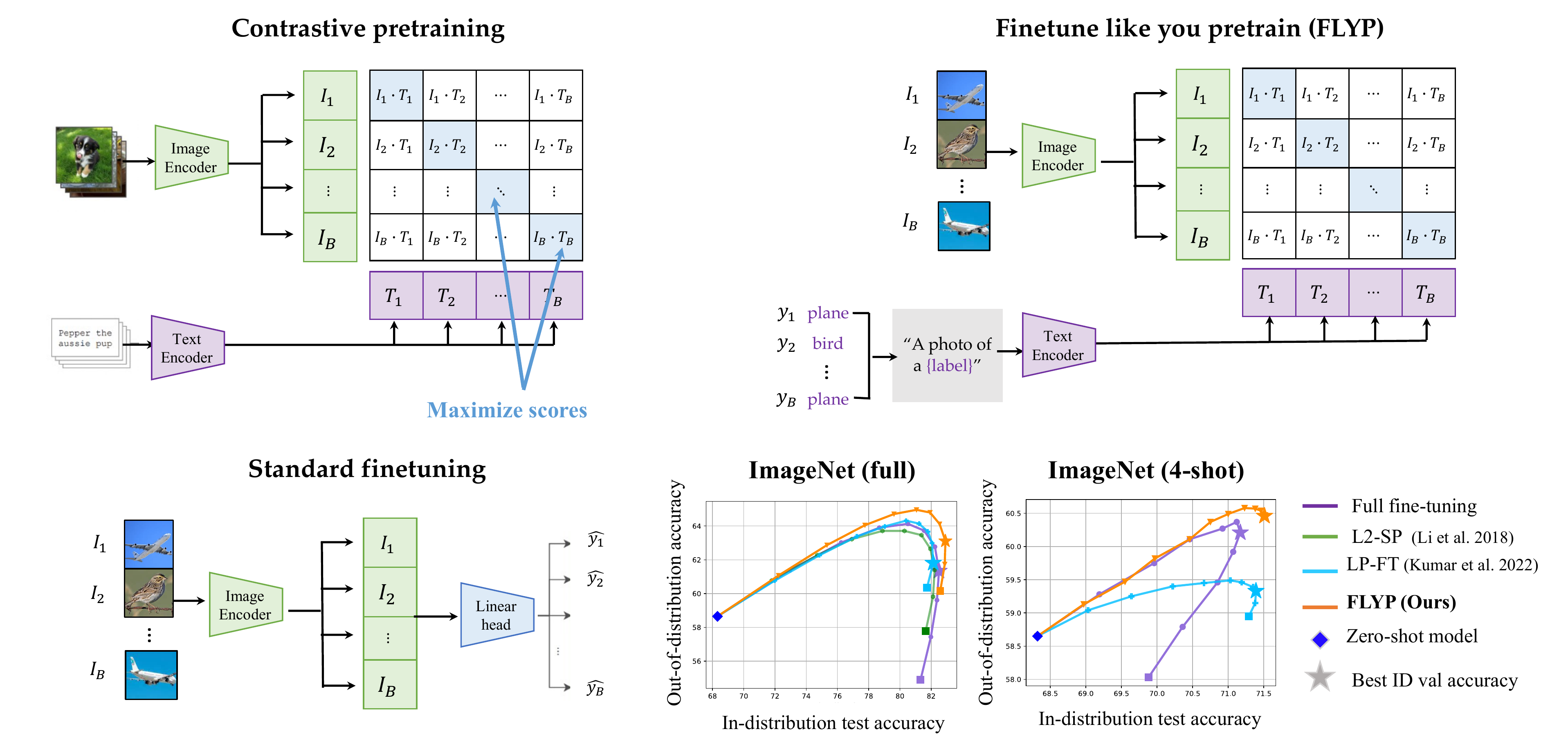}
    \caption{Finetune Like You Pretrain (\method{}): Given a downstream classification dataset, standard finetuning approaches revolve around using the cross-entropy loss. In this work, we show that simply using the same loss as the pretraining i.e. contrastive loss, with ``task supervision'' coming from the text-description of labels, consistently outperforms state-of-the-art approaches like LP-FT~\citep{lpft} and WiseFT~\citep{wiseft}. For example, on ImageNet, our proposed approach outperforms LP-FT + weight ensembling by $1.1\%$ ID and $1.3\%$ OOD, with a ID-OOD frontier curve (orange curve) dominating those of the baselines, i.e. lies above and to the right of all the baselines.}
    \label{fig:intro_fig}
\end{figure*} 

\looseness=-1
A common theme in all these previous methods is that they are small changes to the standard \emph{supervised} training paradigm where we minimize a cross-entropy loss on an image classifier. Indeed, such a choice is natural precisely because we are finetuning the system to improve classification performance. However, directly applying the supervised learning methodology for finetuning pretrained models without considering pretraining process can be sub-optimal. 

In this paper, we show that an alternative, straightforward approach reliably outperforms these previous methods. Specifically, we show that simply finetuning a classifier via the \emph{same pretraining (contrastive) loss} leads to uniformly better performance of the resulting classifiers. That is, after constructing 
prompts from the class labels, we directly minimize the contrastive loss between these prompts and the image embeddings of our (labeled) finetuning set. We call this approach \emph{finetune like you pretrain}  (\method) and summarize in Figure~\ref{fig:intro_fig}. \method results in better ID \emph{and} OOD performance than alternative approaches without any additional features such as multi-stage finetuning or ensembling. When ensembling, it further boosts gains over ensembling with previous methods. This contrastive finetuning is done entirely ``naively'': it ignores the fact that classes within a minibatch may overlap or that multiple prompts can correspond to the same class. 

We show that on a variety of different models and tasks, this simple \method approach consistently outperforms the existing state-of-the-art finetuning methods.  On WILDS-iWILDCam, FLYP gives the highest ever reported accuracy, outperforming the top of the leaderboard (compute expensive ModelSoups~\citep{wortsman2022modelsoups} which ensembles over 70+ finetuned models) by $2.3\%$ ID and $2.7\%$ OOD, using CLIP ViT-L/14@336px. On CLIP ViT-B/16, averaged across $7$ out-of-distribution (OOD) datasets (2 WILDS and 5 ImageNet associated shifts), \method gives gains of $4.2\%$ OOD over full finetuning and of more than $1\%$ both ID and OOD over the \emph{current state of the art}, LP-FT.  We also show that this advantage holds for few-shot finetuning, where only a very small number of examples from each class are present. For example, on binary few-shot classification, our proposed approach \method outperforms the baselines by $4.4\%$ on Rendered-SST2 and $3.8\%$ on PatchCamelyon dataset.
Arguably, these few-shot tasks represent the most likely use case for zero-shot finetuning, where one has both an initial prompt, a handful of examples of each class type, and wishes to build the best classifier possible. 


The empirical gains of our method are quite intriguing. We discuss in Section~\ref{sec:discussion} how several natural explanations and intuitions from prior work fail to explain \emph{why} the pretraining loss works so well as a finetuning objective. For example, one could hypothesize that the gains for FLYP come from using the structure in prompts or updating the language encoder parameters. However, using the same prompts and updating the image and language encoders, but via a cross-entropy loss instead performs worse than FLYP. Furthermore, when we attempt to correct for the overlap in classes across a minibatch, we surprisingly find that this decreases performance. This highlights an apparent but poorly understood benefit to finetuning models on the \emph{same} loss which they were trained upon, a connection that has been observed in other settings as well \citep{tta}.

We emphasize heavily that the contribution of this work does not lie in the novelty of the \method finetuning procedure itself: as it uses the \emph{exact same} contrastive loss as used for training, many other finetuning approaches have used slight variations of this approach (see Section \ref{sec:related_works} for a full discussion of related work).  Rather, the contribution of this paper lies precisely in showing that this extremely naive method, in fact, \emph{outperforms} existing (and far more complex) finetuning methods that have been proposed in the literature. While the method is simple, the gains are extremely surprising, presenting an interesting avenue for investigating the finetuning process. 
In total, these results point towards a simple and effective approach that we believe should be adopted as the ``standard'' method for finetuning zero-shot classifiers rather than tuning via a traditional supervised loss.
\section{Preliminaries}\label{sec:preliminary}
\paragraph{Task.} Consider an image classification setting where the goal is to map an image $I \in \sI$ to a label $y \in \sY$. We use image-text pretrained models like CLIP that learn joint embeddings of image and text. Let $\fimg: \sI \mapsto \R^d$ denote the image encoder that maps an image to a $d-$dimensional image-text embedding space. $\fimg$ is parameterized by parameters $\thetaimg$. Let $\sT$ be the space for text descriptions of images. Analogously, $\ftext: \sT \mapsto \R^d$ is the language encoder with model parameters $\thetatext$.  

\paragraph{Contrastive pretraining with language supervision.} The backbone of the pretraining objective is \emph{contrastive learning}, where the goal is to align the embedding $\fimg(I_i)$ of an image close to the embedding $\ftext(T_i)$ of its corresponding text description, and away from other text embeddings $\ftext(T_j)$ in the batch. Given a batch with $B$ images with their corresponding text descriptions $D = \{ (I_1, T_1), \hdots (I_B, T_B) \}$, pretraining objective is as follows: 
\begin{align}
\label{eq:contrastive_loss}
    \lpretrain (D, \theta):= &\sum \limits_{i=1}^B -\log \frac{\exp\big({\nfimg}(I_i) \cdot {\nftext}(T_i)\big)}{\sum_{j=1}^B \exp\big({\nfimg}(I_i) \cdot {\nftext}(T_j)\big)} ~ + \nonumber \\
    &\sum \limits_{i=1}^B -\log \frac{\exp\big({\nfimg}(I_i) \cdot {\nftext}(T_i)\big)}{\sum_{j=1}^B \exp\big({\nfimg}(I_j) \cdot {\nftext}(T_i)\big)}, 
\end{align}
where $\theta = [\thetaimg, \thetatext]$ are image and text encoder parameters, and $\nfimg$ and $\nftext$ are the $\ell_2$ normalized versions of $\fimg$ and $\ftext$ respectively . CLIP~\citep{clip} uses the same pretraining objective, which can be viewed as training on a proxy classification task consisting of $1$ image and $B$ classes from text embeddings (and symmetrically $1$ text and $B$ classes from image embeddings). 
\vspace{-1em}
\paragraph{Finetuning pretrained models.} We are given access to a few training samples $\{ (x_1, y_2), \hdots (x_n, y_n) \} \sim \Pid$, corresponding to the downstream image classification task of interest. Standard methods of leveraging pretrained image-text models like CLIP are as follows. 
\begin{enumerate}
    \itemsep0em 
    \item Zero-shot (ZS): Since the pretrained image embeddings are trained to be aligned with the text embeddings, we can perform zero-shot classification without updating any weights. Given $k$ classes (names) $\{c_1, c_2, \hdots c_k \}$, we construct corresponding text descriptions $\{T_1, \hdots T_k \}$ using templates (for e.g. ``a photo of a {$c_i$}''). The zero-shot prediction corresponding to image $I$ is $\argmax_i \nftext(T_i)^\top \nfimg(I)$, where $\nftext$ and $\nfimg$ are the normalized text and image embeddings. This can be written as $\argmax_i (\zshead^\top \nfimg(I))_i$ where $\zshead \in \R^{d \times k}$ is the zero-shot linear head with columns corresponding to text descriptions of the classes $T_k$. It is also typical to use multiple templates and sample the text prompt from some  $p_\text{text}(\cdot \mid y)$ and ensemble predictions over multiple prompts.  
    \item Linear probing (LP): We learn a linear classifier $\hclass \in \R^{d \times k}$ on top of frozen image embeddings $\nfimg(I)$ by minimizing the cross-entropy on labeled data from the downstream distribution. 
    \item Full finetuning (FFT): In full finetuning, we update both a linear head $\hclass \in \R^{d \times k}$ and the parameters of the image encoder $\thetaimg$ (initialized at the pretrained value) by minimizing the cross-entropy loss on labeled downstream data. Rather than initializing randomly, we use the zero-shot weights $\zshead$ to initialize the linear head, similar to ~\citet{wiseft}.
    \item LP-FT~\citep{lpft}: Here, we perform a two-stage finetuning process where we first perform linear probing and then full finetuning with $\hclass$ initialized at the linear-probing solution obtained in the first stage. 
    \item Weight-ensembling~\citep{wiseft}: We ensemble the weights by linearly interpolating between the weights of the zero-shot model and a finetuned model. Let $\thetaimg$ denote the pretrained weights of the image encoder, and $\thetaimg'$ denote the finetuned weights. Then weights of weight ensembled model are given as:
\begin{equation}
\label{eq:weight_ensemble}
    \theta_\text{we} = \alpha\thetaimg' + (1-\alpha)\thetaimg, \text{ where } \alpha\in [0,1]
\end{equation}
\end{enumerate}

\section{\ours: Finetune like you pretrain}\label{sec:method}
\looseness=-1
Recall that we are interested in using a pretrained model like CLIP to improve performance on a classification task given access to labeled data corresponding to the task. In this section, we describe our method \ours~which is essentially to continue pretraining, with ``task supervision'' coming from text descriptions of the corresponding target class names in the dataset. The algorithm is the most natural extension of pretraining. 

\looseness=-1
Given a label $y$, let $\sT_y$ denote a set of possible text descriptions of the class. Let $P_\text{text}(\cdot \mid y)$ denote the uniform distribution over possible text descriptions. For example, these descriptions could include different contexts such as ``a photo of a small \{class\}'' as considered in~\citep{clip}.

Given a batch of labeled samples $D = \{(I_1, y_1), \hdots (I_B, y_B) \}$, we construct a corresponding batch $D'$ of image-text pairs and update the model parameters via stochastic gradient descent on the \emph{same pretraining objective} (Equation~\ref{eq:contrastive_loss}). We summarize this in Algorithm~\ref{alg:contrastive_finetuning}.

Inference using the finetuned encoders $\fimg'$ and $\ftext'$ is performed in the same way as zero-shot prediction, except using the finetuned encoders $\fimg'$ and $\ftext'$. Precisely, the prediction for an image I is again given by $\argmax_i \nftext'(T_i)^\top \nfimg'(I)$.
\begin{algorithm}[]
  \caption{\method: Contrastive Finetuning (One batch)}
  \label{alg:contrastive_finetuning}
    \hspace*{\algorithmicindent}
    \textbf{Given:} Pretrained parameters $\thetaimg$ and $\thetatext$,  \hspace*{\algorithmicindent} \hspace*{\algorithmicindent} \hspace*{\algorithmicindent} 
    
    \hspace*{\algorithmicindent} \hspace*{\algorithmicindent} \hspace*{\algorithmicindent} Labeled batch~$D = \{(x_1, y_1), \hdots (x_B, y_B) \}$,  \\
    \hspace*{\algorithmicindent} \hspace*{\algorithmicindent} \hspace*{\algorithmicindent} 
     Distribution over text descriptions $P_\text{text}(\cdot \mid y), y \in \sY$ \\
     \hspace*{\algorithmicindent} \hspace*{\algorithmicindent} \hspace*{\algorithmicindent}  
     Learning rate $\alpha$ \\
    \hspace*{\algorithmicindent} \textbf{Training step:}\\
        \hspace*{\algorithmicindent} \hspace*{\algorithmicindent} \hspace*{\algorithmicindent} 1. Create text/image paired data via labels
        \[    D' = \{ (I_1, T_1), \hdots (I_B, T_B) \},  ~\text{where}~T_i \sim P_\text{text}(\cdot \mid y_i) \]
        \hspace*{\algorithmicindent} \hspace*{\algorithmicindent} \hspace*{\algorithmicindent} 2. Update parameters via contrastive loss
        \[ \theta := \theta - \alpha \nabla \lpretrain(D', \theta) \;(\mbox{Equation}~\ref{eq:contrastive_loss}). \quad\quad\quad\;\; \]
\end{algorithm}
\vspace{-1em}
\paragraph{\ours~ vs. standard finetuning.} 
The finetuning loss presented above is the most natural extension of the pretraining objective to incorporate labeled downstream data. However, this differs from what is currently the standard practice for finetuning CLIP models. We remark on the main differences (details in Section~\ref{sec:discussion}) to determine the effect of various contributing factors. 

(1) FLYP updates the language encoders: Standard finetuning methods typically only update the image encoder, while FLYP essentially continues pretraining which updates both the image and language encoders. However, we show in Section~\ref{sec:discussion} that FLYP's gains are not simply from updating the language encoder---using the cross-entropy loss but updating the language encoder gets lower accuracy than FLYP. 

(2) The linear head in FLYP incorporates structure from the text embeddings of corresponding labels (e.g., some classes are closer than others). We simulate this effect for cross-entropy finetuning by using the embeddings of text-description of classnames as the linear head (detailed in Section~\ref{sec:discussion}). However, we find that it still performs worse than FLYP, highlighting that the choice of loss function matters indeed.

In summary, \method includes some intuitively favorable factors over standard finetuning, but via our ablations, these do not fully explain the success of \method. It appears that fine-tuning in exactly the same way as we pretrain is important for \method's success.

\section{Experiments}\label{sec:experiment}

\method outperforms other finetuning methods on 8 standard datasets across 3 settings. We show results for distribution shift benchmarks, which was the focus of the original CLIP paper~\citep{clip} and recent finetuning innovations~\citep{lpft}, in Section~\ref{sec:eval_dist_shift}.
We then show results for few-shot learning in Section~\ref{sec:eval_few_shot}, and standard transfer learning tasks in Section~\ref{sec:eval_transfer}. Finally in Section~\ref{sec:discussion}, we talk about various possible explanations to effectiveness of \method. We find that finetuning the language encoder, and using the pretraining loss instead of a standard cross-entropy loss, are both critical to \method's success.
\vspace{-1em}
\paragraph{Datasets:} 
\label{sec:datasets}
\begin{enumerate}
    \itemsep0em 
    \item \textbf{ImageNet}~\citep{russakovsky2015imagenet} is a large-scale dataset containing over a million images, where the goal is to classify an image into one of 1000 categories. We finetune on ImageNet as the ID dataset and evaluate all five standard OOD datasets considered by prior work~\citep{radford2021clip,wortsman2021robust,kumar2022finetuning}: \textbf{ImageNetV2}~\citep{recht2019doimagenet}, \textbf{ImageNet-R}~\citep{hendrycks2020many}, \textbf{ImageNet-A}~\citep{hendrycks2019natural}, \textbf{ImageNet-Sketch}~\citep{wang2019learningrobust}, and \textbf{ObjectNet}~\citep{barbu2019objectnet}.
    \item \textbf{WILDS-iWILDCam}~\citep{beery2020iwildcam, koh2021wilds} is a 182 class classification dataset of animal images. The ID and OOD datasets differ in the camera used and  factors like background, illumination, etc.
    \item \textbf{WILDS-FMoW}~\citep{christie2018fmow, koh2021wilds} consists of remote sensing satellite images. The goal is to classify a satellite image into one of 62  categories, such as ``impoverished settlement'' or ``hospital''. The ID and OOD datasets differ in the time of their collection and location, i.e., continent.
    \itemsep0em 
    \item \textbf{Caltech101}~\citep{caltech101_dataset} has pictures of various objects belonging to 101 categories like ``helicopter'', ``umbrella'', ``watch'', etc.
    \item \textbf{StanfordCars}~\citep{stanfordcars_dataset} consists of images of cars with different models, make, and years. The goal is to classify into one of 196 categories, like ``Audi 100 Sedan 1994'' or "Hyundai Sonata Sedan 2012". 
    \item \textbf{Flowers102}~\citep{flowers102_dataset} has images of flowers occurring in UK. The goal is  to classify into one of 102 categories like ``fire lily'' or ``hibiscus''.
    \item \textbf{PatchCamelyon}~\citep{pcam_datasets} is a challenging binary classification dataset consisting of digital pathology images. The goal is to detect the presence of metastatic tumor tissues.
    \item \textbf{Rendered SST2}~\citep{clip} is an optical character recognition dataset, where the goal is to classify the text sentiment into ``positive'' or ``negative''.
\end{enumerate}
\vspace{-15pt}
\paragraph{Baselines.} We compare with two most standard ways of adapting pretrained models: linear probing (LP) and end-to-end full finetuning (FFT), using the cross-entropy loss. We also compare with recently proposed improvements to finetuning: L2-SP \citep{l2sp} where the finetuned weights are regularized towards the pretrained weights, and LP-FT~\citep{lpft} where we first do linear probe followed by full finetuning. Additionally, we compare all methods with weight ensembling as proposed in WiseFT~\citep{wiseft}.

\vspace{-1em}
\paragraph{Models.}We consider three models: CLIP ViT-B/16 and a larger CLIP ViT-L/14 from OpenAI~\citep{clip}, and a CLIP ViT-B/16 trained on a different pretraining dataset (public LAION dataset~\citep{laion400m})  from~\citet{openCLIP}. The default model used is the CLIP ViT-B/16 from OpenAI unless specified otherwise.
\vspace{-1em}
\paragraph{Experiment protocol.} For all baselines and our proposed approach, we finetune with AdamW using a cosine learning rate scheduler. We use a batch-size of 512 for ImageNet and 256 for all other datasets. We sweep over learning rate and weight decay parameters and early stop all methods using accuracy on the in-distribution (ID) validation set. OOD datasets are used only for evaluation and not for hyperparameter sweeps or early stopping. We report $95\%$ confidence intervals over $5$ runs. For few-shot learning, we use a validation set that is also few-shot and report accuracy over $50$ repeated runs due to increased variance caused by a small validation set. For all the datasets, we use the same text-templates as used in CLIP~\citep{clip} and WiseFT~\citep{wiseft}. For details on hyperparameter sweeps, see Appendix B.




\subsection{Evaluation Under Distribution Shifts}
\label{sec:eval_dist_shift}
\begin{figure*}[htb]
    \centering 
\begin{subfigure}{0.32\textwidth}
  \includegraphics[width=\linewidth]{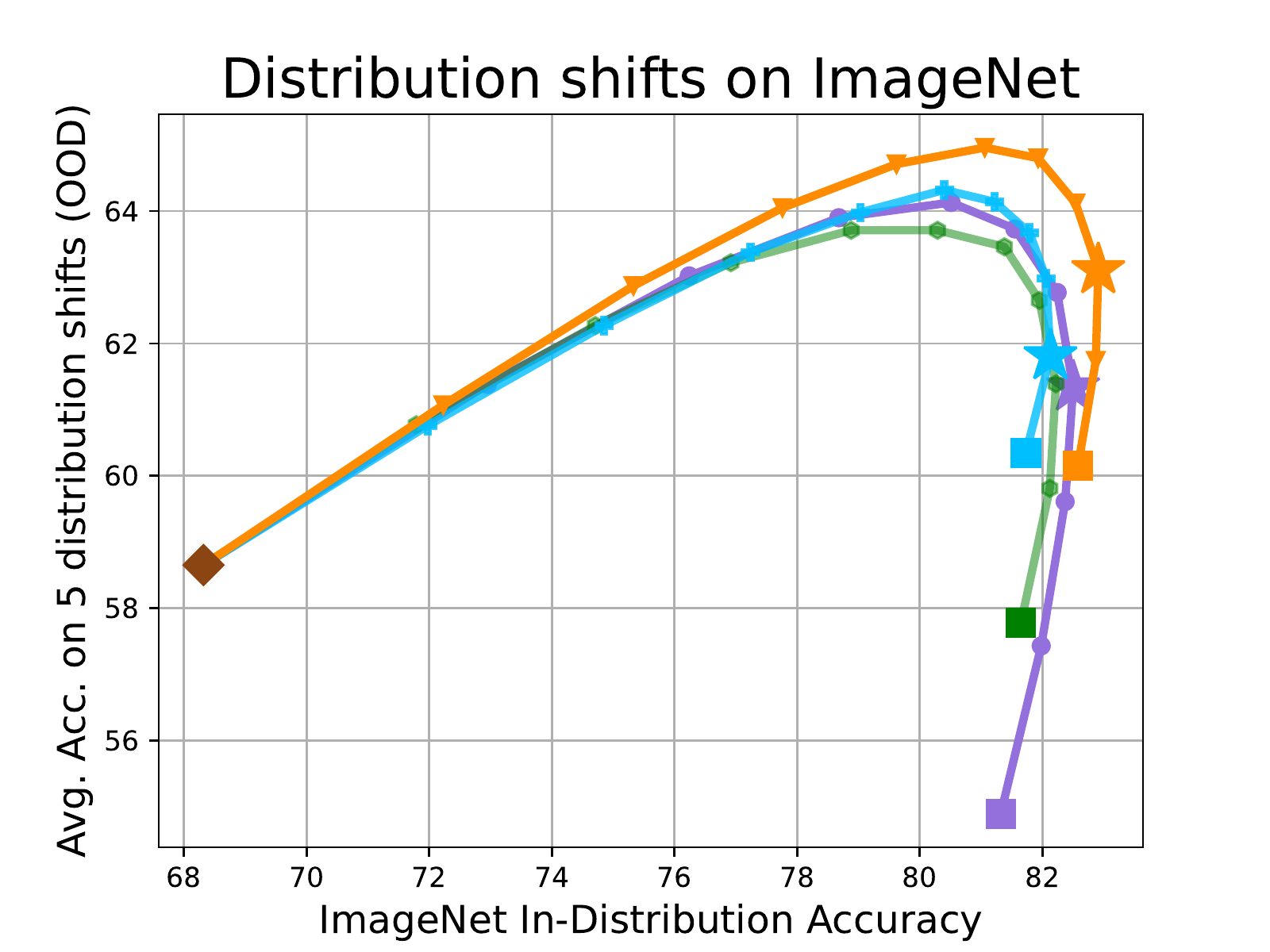}
  \caption{\centering OpenAI CLIP ViT-B/16 finetuned on ImageNet}
  \label{fig:distribution_shifts_imagenet}
\end{subfigure}\hfil 
\begin{subfigure}{0.32\textwidth}
  \includegraphics[width=\linewidth]{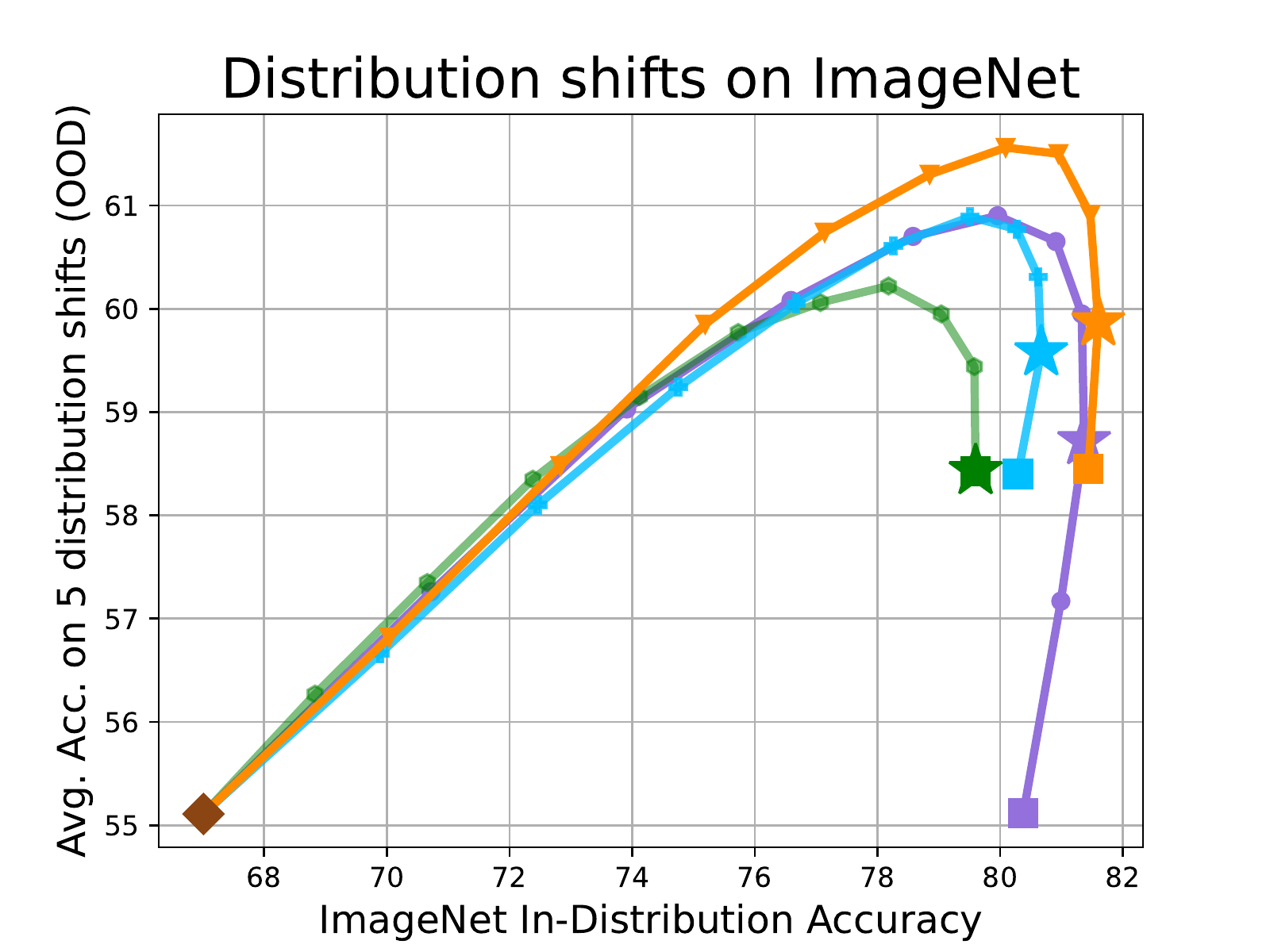}
  \caption{\centering LAION-400M CLIP ViT-B/16, finetuned on ImageNet}
  \label{fig:distribution_shifts_imagenet_laion}
\end{subfigure}\hfil 
\begin{subfigure}{0.32\textwidth}
  \includegraphics[width=\linewidth]{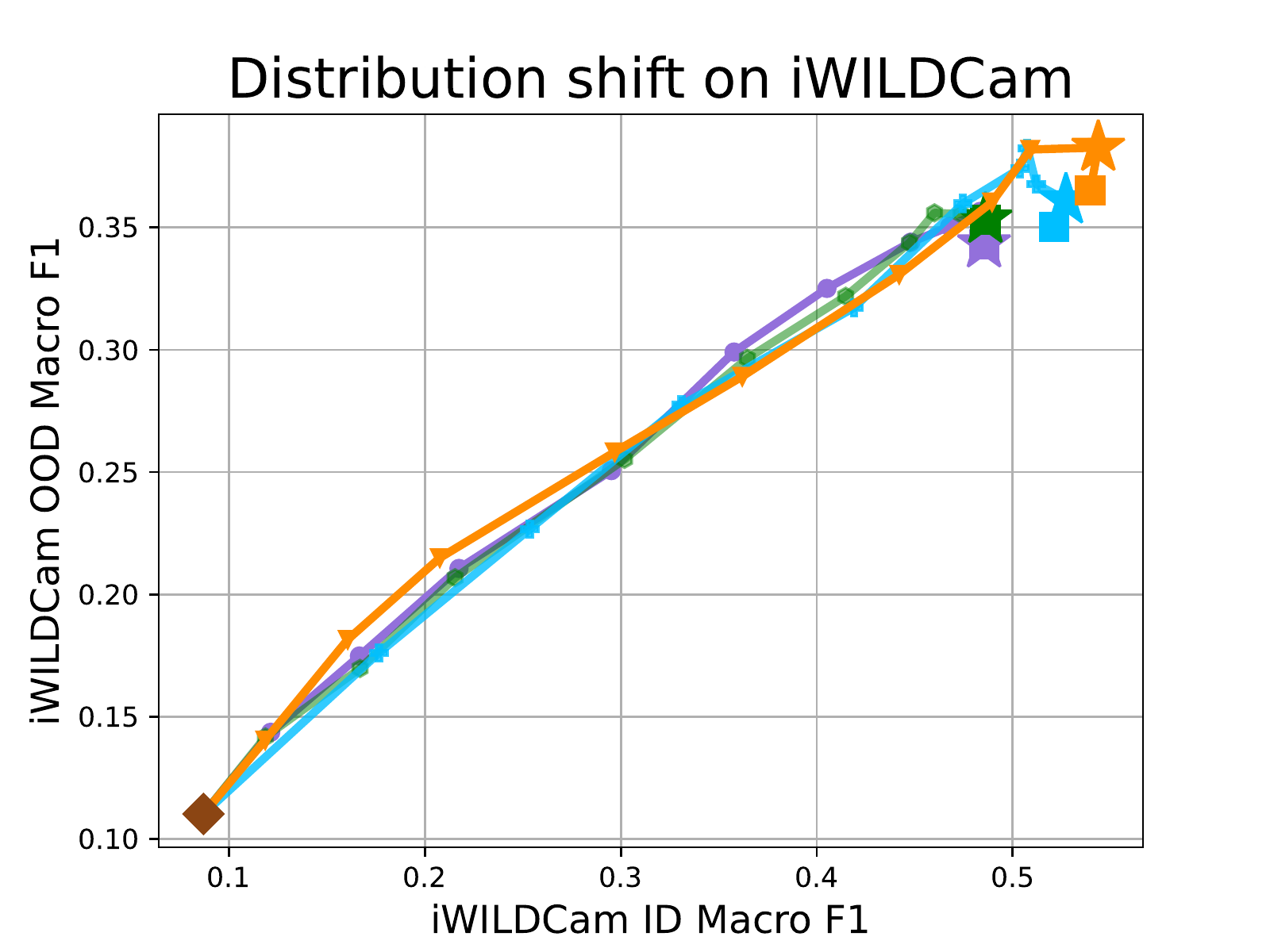}
  \caption{\centering CLIP ViT-B/16 finetuned on WILDS-iWILDCam}
  \label{fig:distribution_shifts_iwildcam}
\end{subfigure}\hfil 
\medskip
\begin{subfigure}{0.32\textwidth}
  \includegraphics[width=\linewidth]{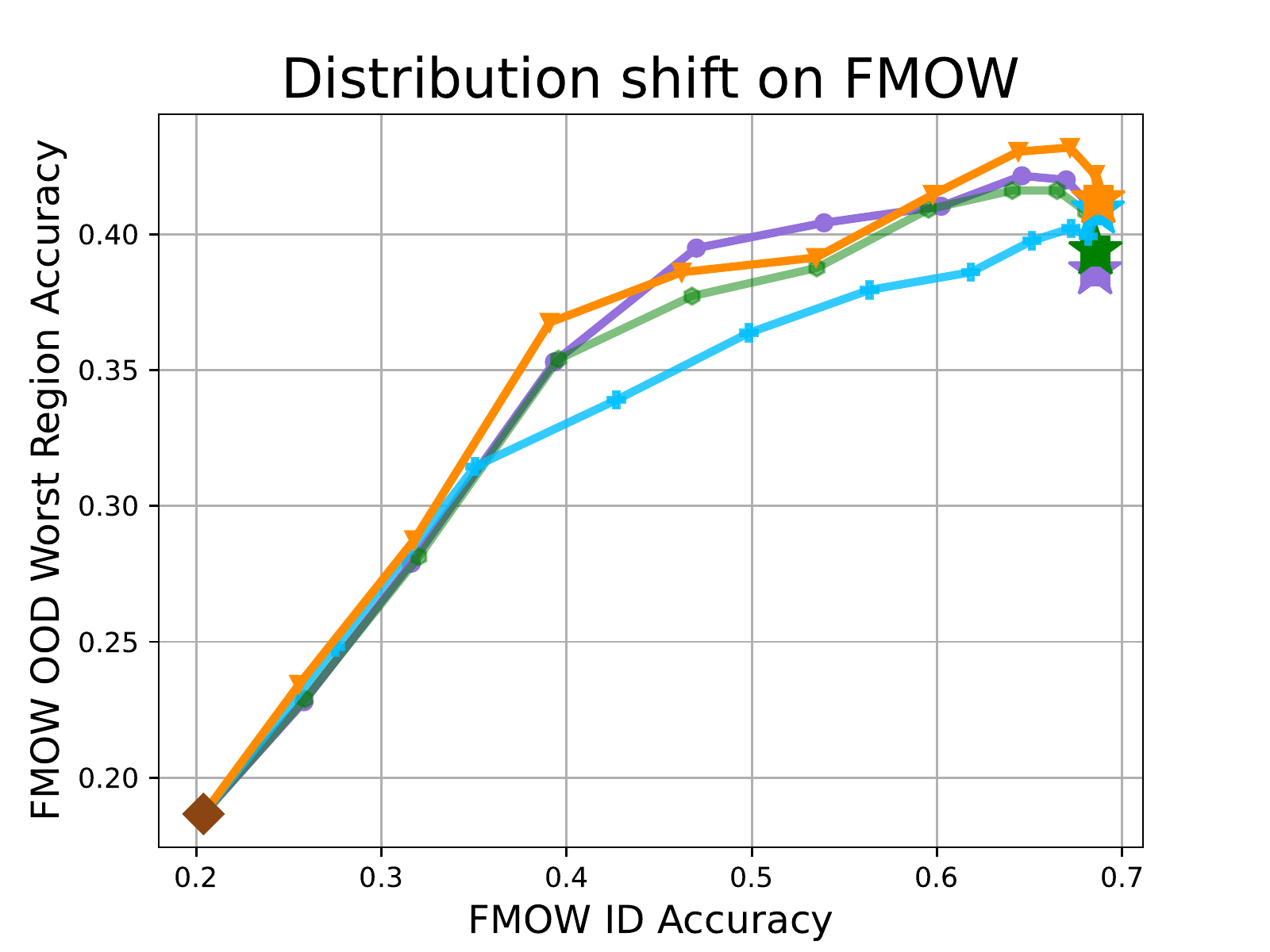}
  \caption{\centering CLIP ViT-B/16 finetuned on WILDS-FMoW}
  \label{fig:distribution_shifts_fmow}
\end{subfigure}\hfil 
\begin{subfigure}{0.32\textwidth}
  \includegraphics[width=\linewidth]{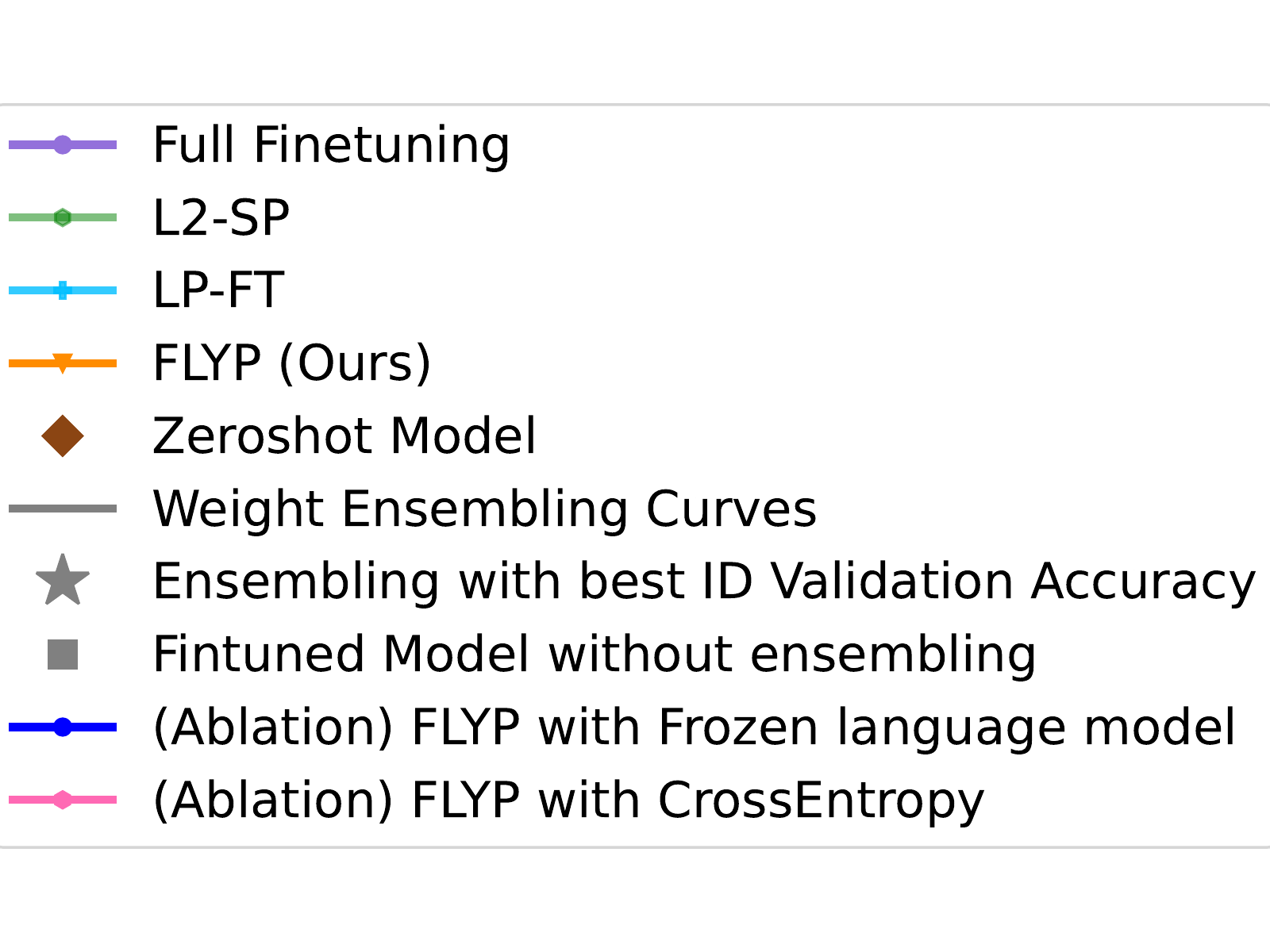}
  \caption{\centering Legend for all the subplots}
  \label{fig:distribution_shifts_legend}
\end{subfigure}\hfil 
\begin{subfigure}{0.32\textwidth}
  \includegraphics[width=\linewidth]{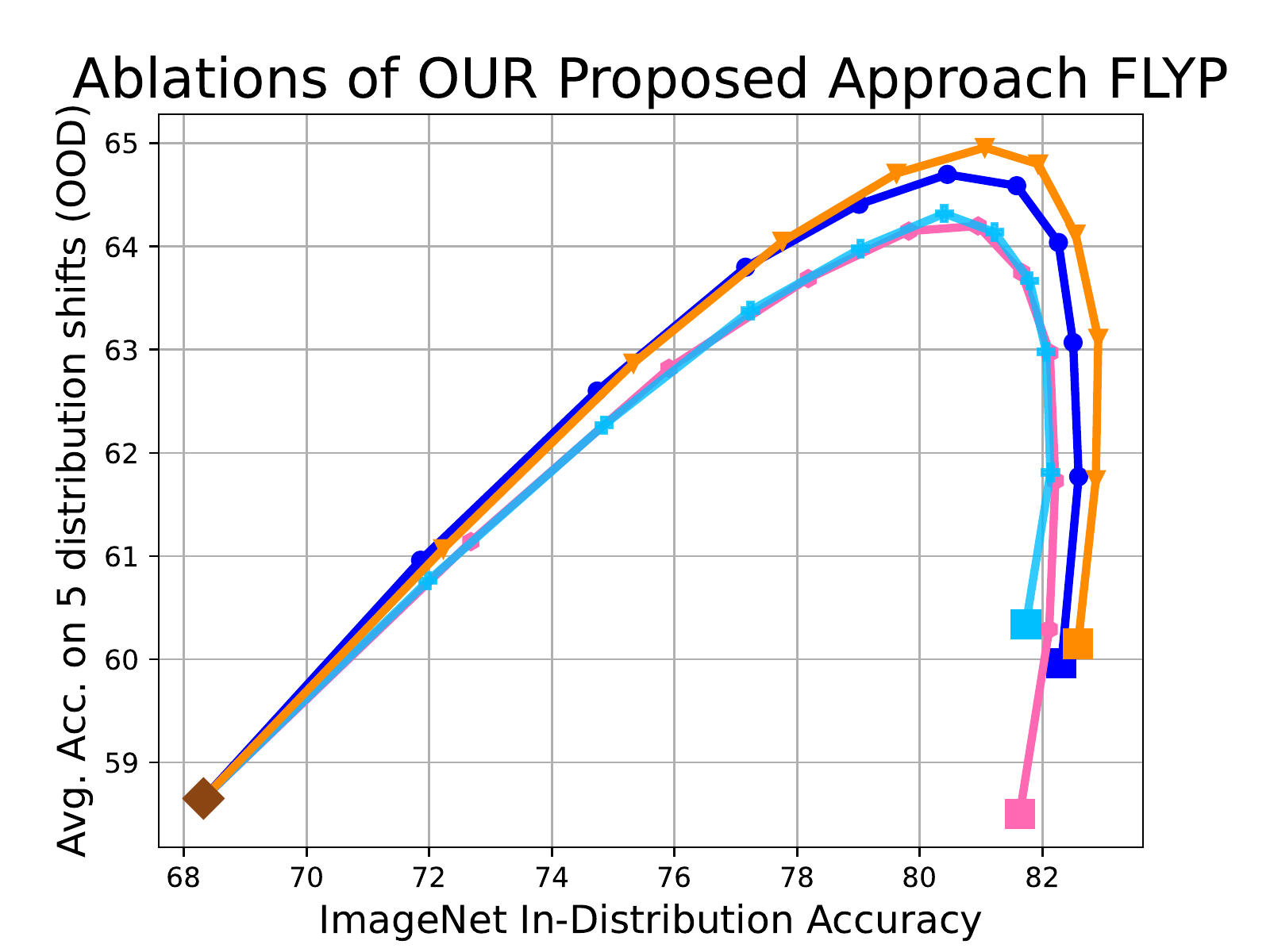}
  \caption{Ablation of \method, details in Section~\ref{sec:discussion}}
  \label{fig:ablation}
\end{subfigure}
\caption{Our proposed approach \method outperforms the baselines both ID and OOD, with or without weight ensembling~\citep{wiseft}. Here we show the ID-OOD frontier curves obtained by linearly interpolating the finetuned model weights with the zeroshot weights. The curves for \method completely dominate (lies above and to the right) those of the baselines on ImageNet, giving higher OOD accuracy for any ID accuracy. Comparing with ensembling corresponding to the best ID validation accuracy(stars), \method outperforms the current state of the art, LP-FT, by an average of $1.3\%$ OOD and $1.1\%$ ID and outperforms WiseFT (weight ensembled finetuning, ~\cite{wiseft}) by an average of $2\%$ OOD and $1.6\%$ ID. We report exact numbers in Table~\ref{tab:big_distribution_shift_table}.}
\label{fig:distribution_shifts} 
\vspace{-1em}
\end{figure*}

\method outperforms baselines on ImageNet, WILDS-iWildCam, WILDS-FMoW, and associated distribution shifts both ID and OOD. Infact, \method outperforms the highest previously reported accuracy on iWILDCam by $2.3\%$ ID and $2.7\%$ OOD, using ViT-L/14@336px. On ViT-B/16, averaged across $7$ out-of-distribution (OOD) datasets, \method gives gains of $4.2\%$ over full finetuning and $1.3\%$ with weight ensembling over the \emph{current state of the art}, LP-FT. Note that these gains in OOD accuracy do \emph{not} come at the cost of ID accuracy--- averaged over the same datasets, \method outperforms full finetuning by $1.8\%$ ID and LP-FT by $1.2\%$ ID. 
\vspace{-1em}
\paragraph{Weight ensembling curves:} WiSE-FT~\citep{wiseft} shows that a simple linear interpolation between the weights of the pretrained and the finetuned model gives the best of both in-distribution (ID) and out-of-distribution(OOD) performance. This gain comes at no training cost and importantly, allows to trace the whole ID-OOD accuracy frontier and check if the OOD gains are at the cost of ID accuracy. Hence for our main results we compare the baselines and \method while interpolating their weights with 10 mixing coefficients $\alpha\in[0,1]$ (Eqation~\ref{eq:weight_ensemble}). The resultant ID-OOD ``frontier'' curves are then evaluated by comparing their ID-OOD accuracy at the choice of coefficient which gives the highest ID validation accuracy. 
\vspace{-1em}
\paragraph{Evaluation:} In Table~\ref{tab:big_distribution_shift_table}, for all the baselines and \method, we report the ID-OOD accuracy with and without weight ensembling at the mixing coefficient having the highest ID validation accuracy. We observe that \method outperforms the baselines with and without weight ensembling. Following the literature~\citep{accuracy_on_line, wiseft}, we use macro F1 score as the ID and OOD evaluation metrics for iWILDCam dataset. For FMOW, we report worst region OOD accuracy.
\vspace{-1em}
\paragraph{Improves accuracy on ImageNet and shifts.} We show that for any ID accuracy, \method obtains better OOD accuracy than baselines. In particular, Figure~\ref{fig:distribution_shifts_imagenet} compares \method (orange curve) with various baselines on ImageNet---we plot the average OOD accuracy on $5$ ImageNet-shift benchmarks against the ID accuracy on ImageNet. The weight ensembling curve for \method dominates (lies entirely above and to the right) those of the baselines. When choosing the mixing coefficient which gives the best ID validation accuracy (stars on the respective curves and Table~\ref{tab:big_distribution_shift_table}), \method outperforms WiseFT(weight ensembled finetuning) by $2\%$ OOD and the \emph{current state of the art}, LP-FT, by $1.4\%$ OOD. 

Even without weight ensembling, as shown in Table~\ref{tab:big_distribution_shift_table}, \method outperforms full finetuning by $1.2\%$ ID and $5.4\%$ OOD and similarly gives ID accuracy gains over LP-FT with almost the same OOD accuracy. 

Figure~\ref{fig:distribution_shifts_imagenet_laion} shows that the same observations hold when we use a ViT-B/16 from~\citet{openCLIP} (trained on a different dataset). Here, \method slightly outperforms LP-FT on OOD even without weight ensembling. In Appendix A we give results for all the 5 OOD datasets, and show that FLYP consistently outperforms on all of them.

\begin{table*}[t!]
\centering
\resizebox{0.95\textwidth}{!}{
\begin{tabular}{@{}ccccccccccc@{}}
\toprule
                          & \multicolumn{4}{c}{Imagenet}                                                                                                       & \multicolumn{4}{c}{iWILDCam}                                                                                                       & \multicolumn{2}{c}{FMoW}                  \\ \cmidrule(l){2-11} 
                          & \multicolumn{2}{c}{Without Ensembling}    & \multicolumn{2}{c|}{With Ensembling}                                                   & \multicolumn{2}{c}{Without Ensembling}    & \multicolumn{2}{c|}{With Ensembling}                                                   & \multicolumn{2}{c}{Without Ensembling}    \\ \cmidrule(l){2-11} 
\multirow{-3}{*}{Methods} & ID                  & OOD                 & ID                  & \multicolumn{1}{c|}{OOD}                                         & ID                  & OOD                 & ID                  & \multicolumn{1}{c|}{OOD}                                         & ID                  & OOD                 \\ \midrule
Zeroshot                  & 68.3 (-)            & 58.7 (-)            & 68.3 (-)            & \multicolumn{1}{c|}{58.7 (-)}                                    & 8.7 (-)             & 11.02 (-)           & 8.7 (-)             & \multicolumn{1}{c|}{11.02 (-)}                                   & 20.4 (-)            & 18.66 (-)           \\
LP                        & 79.9 (0.0)          & 57.2 (0.0)          & 80.0 (0.0)          & \multicolumn{1}{c|}{58.3 (0.0)}                                  & 44.5 (0.6)          & 31.1 (0.4)          & 45.5 (0.6)          & \multicolumn{1}{c|}{31.7 (0.4)}                                  & 48.2 (0.1)          & 30.5 (0.3)          \\
FT                        & 81.4 (0.1)          & 54.8 (0.1)          & 82.5 (0.1)          & \multicolumn{1}{c|}{61.3 (0.1)}                                  & 48.1 (0.5)          & 35.0 (0.5)          & 48.1 (0.5)          & \multicolumn{1}{c|}{35.0 (0.5)}                                  & 68.5 (0.1)          & 39.2 (0.7)          \\
L2-SP                     & 81.6 (0.1)          & 57.9 (0.1)          & 82.2 (0.1)          & \multicolumn{1}{c|}{58.9 (0.1)}                                  & 48.6 (0.4)          & 35.3 (0.3)          & 48.6 (0.4)          & \multicolumn{1}{c|}{35.3 (0.3)}                                  & 68.6 (0.1)          & 39.4 (0.6)          \\
LP-FT                     & 81.8 (0.1)          & \textbf{60.5 (0.1)} & 82.1 (0.1)          & \multicolumn{1}{c|}{61.8 (0.1)}                                  & 49.7 (0.5)          & 34.7 (0.4)          & 50.2 (0.5)          & \multicolumn{1}{c|}{35.7 (0.4)}                                  & 68.4 (0.2)          & 40.4 (1.0)          \\ \midrule
\rowcolor[HTML]{C0C0C0} 
FLYP                      & \textbf{82.6 (0.0)} & 60.2 (0.1)          & \textbf{82.9 (0.0)} & \multicolumn{1}{c|}{\cellcolor[HTML]{C0C0C0}\textbf{63.2 (0.1)}} & \textbf{52.2 (0.6)} & \textbf{35.6 (1.2)} & \textbf{52.5 (0.6)} & \multicolumn{1}{c|}{\cellcolor[HTML]{C0C0C0}\textbf{37.1 (1.2)}} & \textbf{68.6 (0.2)} & \textbf{41.3 (0.8)} \\ \bottomrule
\end{tabular}
}
\caption{\method outperforms the baselines both with and without ensembling. When ensembling, we choose the mixing coefficient (Equation~\ref{eq:weight_ensemble}) with the highest ID validation accuracy. For ImageNet, we report the mean OOD accuracy on $5$ associated distribution shifts, and share individual numbers in appendix. \method outperforms all the baselines in 9 out of 10 various experiment settings. Without weight ensembling, averaged over all the datasets, \method outperforms full finetuning by $4.24\%$ OOD and $1.8\%$ ID. Similarly, \method outperforms LP-FT by $1.2\%$ ID and gives similar OOD performance averaged over all the datasets.}

\label{tab:big_distribution_shift_table}
\end{table*}

\vspace{-1em}
\paragraph{SOTA accuracy on WILDS} 
To verify if the gains using our proposed approach \method can be observed even when using large CLIP models, we evaluate on ViT-L/14@336px. Table~\ref{tab:sota_iwild} compares \method with the leaderboard on iWILDCam benchmark~\citep{wilds_leaderboard}. \method gives gains of $2.3\%$ ID and $2.7\%$ OOD over the top of the leaderboard, outperforming compute heavy ModelSoups~\citep{wortsman2022modelsoups}, which ensemble more than 70+ models trained using various augmentations and hyper-parameters.
Note that ModelSoups uses the state-of-the-art baseline LP-FT for finetuning, which we compare against in this work. Even using the smaller architecture of ViT-B/16, \method outperforms the baselines both with and without ensembling, giving gains of $2.5\%$ ID and and $1\%$ OOD, as shown in Table~\ref{tab:big_distribution_shift_table} and Figure ~\ref{fig:distribution_shifts_iwildcam}.

\renewcommand{\arraystretch}{1}
\begin{table}[t]
\centering
\begin{minipage}[b]{0.5\textwidth}%
\resizebox{1.0\textwidth}{!}{
\begin{tabular}{@{}cccc@{}}
\toprule
\multicolumn{1}{l}{} & Architecture & ID                  & OOD                 \\ \midrule
\rowcolor[HTML]{C0C0C0} 
FLYP                 & ViTL-336px   & \textbf{59.9 (0.7)} & \textbf{46.0 (1.3)} \\
Model Soups          & ViTL         & 57.6 (1.9)          & 43.3 (1.)           \\
ERM                  & ViTL         & 55.8 (1.9)          & 41.4 (0.5)          \\
ERM                  & PNASNet      & 52.8 (1.4)          & 38.5 (0.6)          \\
ABSGD                & ResNet50     & 47.5 (1.6)          & 33.0 (0.6)          \\ \bottomrule
\end{tabular}
}
\end{minipage}
\caption{\method (with ensembling) achieves highest reported accuracy both ID and OOD on WILDS-iWILDCam benchmark. Here we compare \method with the top 4 entries on the leaderboard ~\citep{wilds_leaderboard}.  FLYP uses a slightly different base architecture to achieve the highest result possible, but we note that gains are still present in other experiments using the same base architecture, as shown in Table \ref{tab:big_distribution_shift_table}.}
\label{tab:sota_iwild}
\vspace{-1em}
\end{table}

Similarly, on WILDS-FMOW (Table~\ref{tab:big_distribution_shift_table}), \method outperforms baselines obtaining an OOD accuracy of $41.3\%$ (vs. $40.4\%$ accuracy for LP-FT and $39.4\%$ for L2-SP). Note that we did not observe gains using weight ensembling (when choosing the ensemble with best ID validation accuracy) for any on the baselines.

\subsection{Few-shot classification}
\label{sec:eval_few_shot}
In the challenging setting of few-shot classification,
\method performs impressively well, giving gains as high as $4.4\%$ on SST2 and $3.8\%$ on PatchCamelyon. Even on Imagenet, \method gives a lift of upto $1.5\%$ OOD compared to the most competitive baseline, as we discuss in detail below. Few-shot classification is arguably one of the most likely use case scenarios for zeroshot models, where one has access to a few relevant prompts (to get the zeroshot classifier) as well as a few labeled images, and would want to get the best possible classifier for downstream classification.
\vspace{-10pt}
\subsubsection{Binary few-shot classification}
\label{sec:binary_few_shot}
In few-shot binary classification, the total number of training examples is small, making this a challenging setting. We experiment on $2$ datasets: PatchCamelyon and Rendered-SST2 as used in ~\citet{clip}. \method gives impressive gains across $4, 16 \text{ and }32$ shot classification, as shown in Table~\ref{tab:fewshot_transfer}. For example, on $16$-shot classification on PatchCamelyon, \method outperforms LP-FT by $4.7\%$ and full finetuning by $2.9\%$. On SST2 $32$-shot classification, \method outperforms the most competitive baseline by $4.4\%$. We observe similar accuracy gains when using a much larger CLIP architecture of ViT-L/14 as well, as shown in Appendix A. 

\renewcommand{\arraystretch}{1}
\begin{table}[t!]
\centering
\begin{minipage}[b]{0.75\textwidth}%
\resizebox{1.0\textwidth}{!}{
\begin{tabular}{@{}ccccccc@{}}
\toprule
\multicolumn{1}{l}{} & \multicolumn{3}{c}{PatchCamelyon}                                           & \multicolumn{3}{c}{SST2}                                        \\ \cmidrule(l){2-7} 
k (shots)            & 4                   & 16                  & 32                              & 4                   & 16                  & 32                  \\ \midrule
Zeroshot             & 56.5 (-)            & 56.5 (-)            & \multicolumn{1}{c|}{56.5 (-)}   & 60.5 (-)            & 60.5 (-)            & 60.5 (-)            \\
LP                   & 60.4 (4.0)          & 64.4 (3.7)          & \multicolumn{1}{c|}{67.0 (4.4)} & 60.8 (1.8)          & 61.9 (1.4)          & 62.9 (1.3)          \\
FT                   & 63.1 (5.5)          & 71.6 (4.6)          & \multicolumn{1}{c|}{75.2 (3.7)} & 61.1 (0.7)          & 62.4 (1.6)          & 63.4 (1.9)          \\
LP-FT                & 62.7 (5.3)          & 69.8 (5.3)          & \multicolumn{1}{c|}{73.9 (4.6)} & 60.9 (2.4)          & 62.9 (1.9)          & 63.6 (1.4)          \\ \midrule
\rowcolor[HTML]{C0C0C0} 
FLYP                 & \textbf{66.9 (5.0)} & \textbf{74.5 (2.0)} & \textbf{76.4 (2.4)}             & \textbf{61.3 (2.7)} & \textbf{65.6 (2.1)} & \textbf{68.0 (1.7)} \\ \bottomrule
\end{tabular}
}
\end{minipage}
\caption{In binary few-shot classification, \method performs remarkably well. For example, \method outperforms LP-FT by $4.4\%$ and full finetuning by $4.6\%$ in 32-shot classification on SST2. We observe similar gains when using a much larger CLIP architecture of ViT-L/14, as detailed in Appendix A.}
\label{tab:fewshot_transfer}
\vspace{-1em}
\end{table}

\begin{figure*}[ht!]
    \centering 
\begin{subfigure}{0.25\textwidth}
  \includegraphics[width=\linewidth]{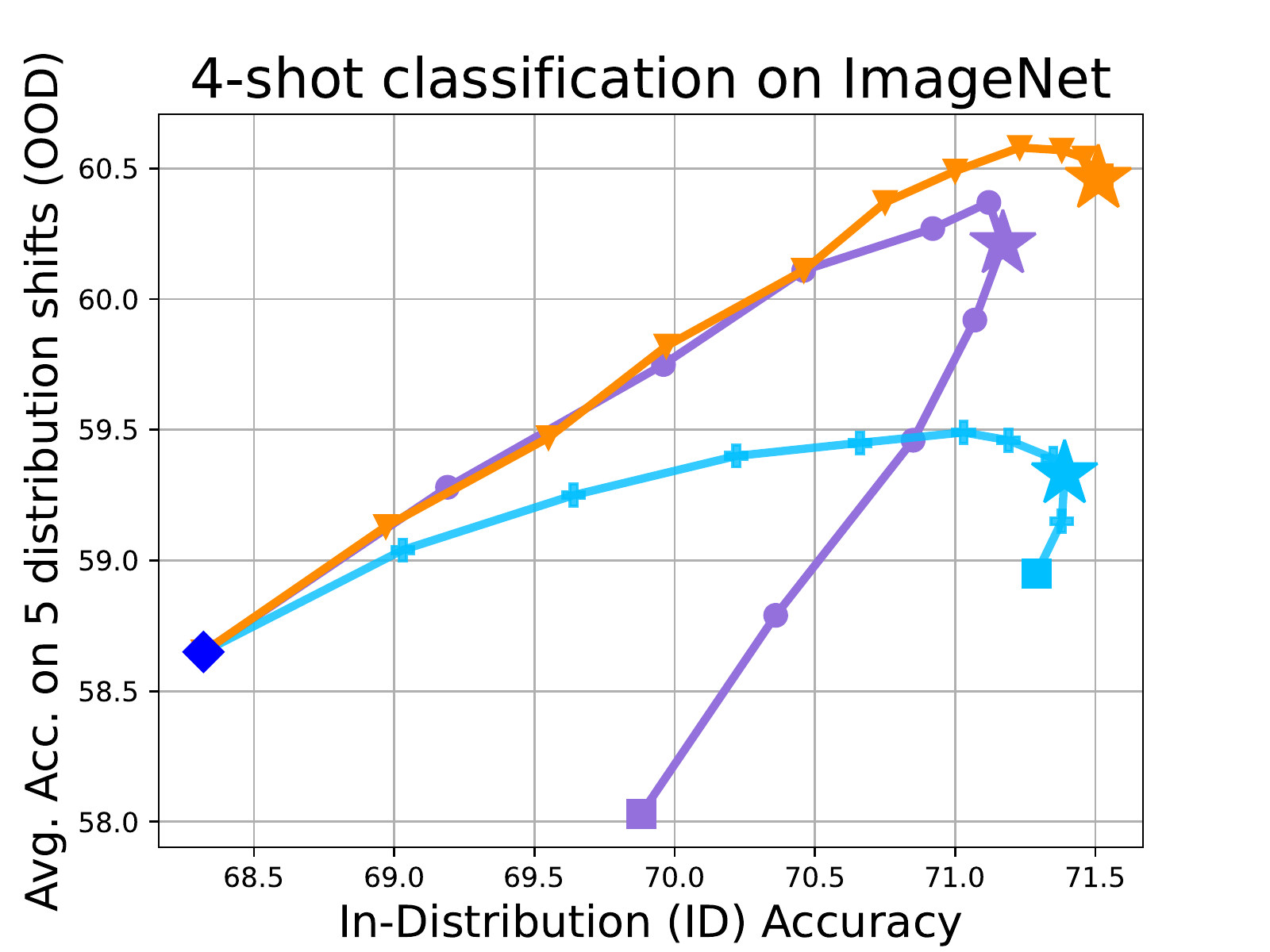}
  \caption{$4$-shot classification}
  \label{fig:fewshot_imnet_4}
\end{subfigure}\hfil 
\begin{subfigure}{0.25\textwidth}
  \includegraphics[width=\linewidth]{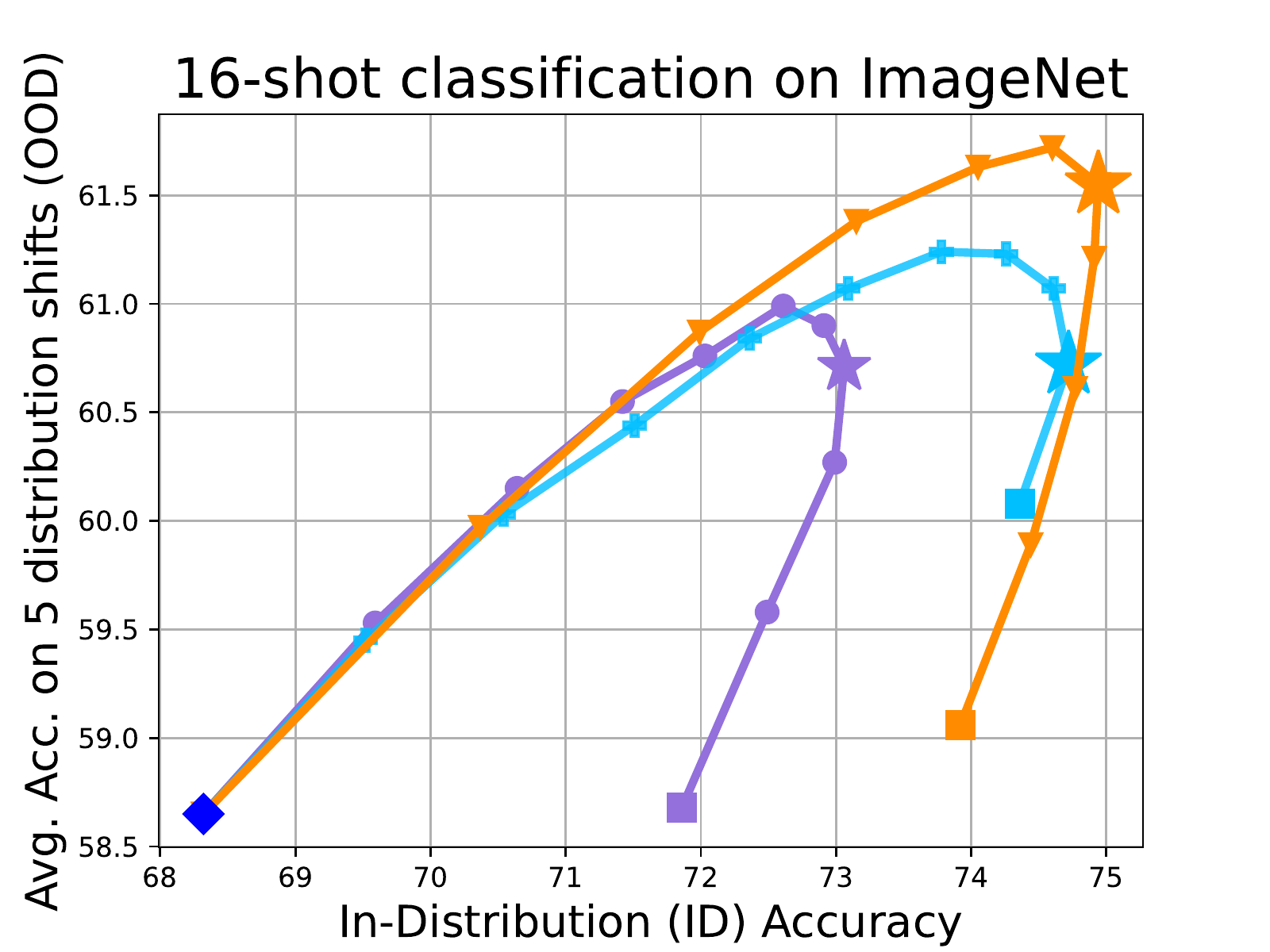}
  \caption{$16$-shot classification}
  \label{fig:fewshot_imnet_16}
\end{subfigure}\hfil 
\begin{subfigure}{0.25\textwidth}
  \includegraphics[width=\linewidth]{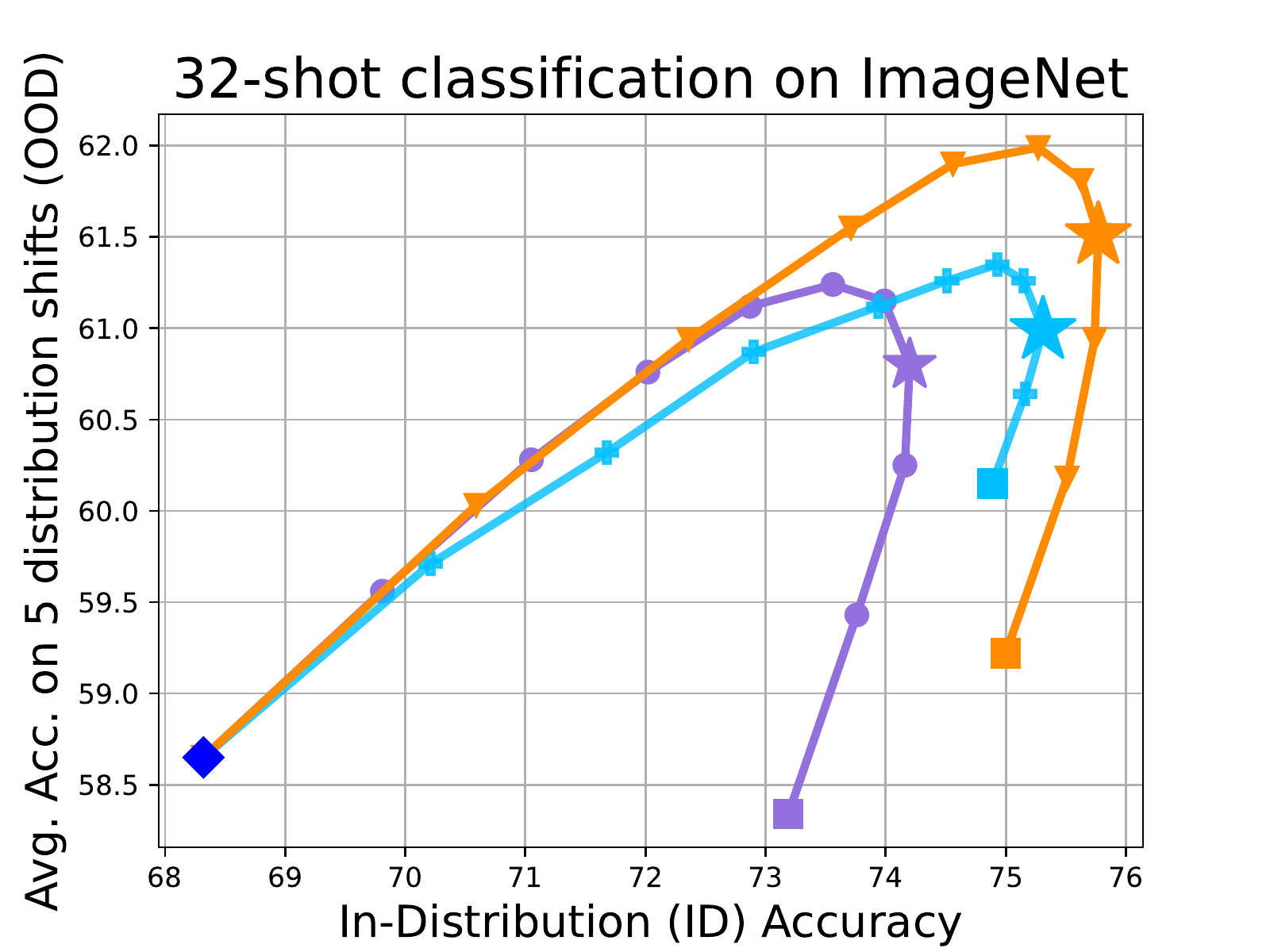}
  \caption{$32$-shot classification}
  \label{fig:legend2}
\end{subfigure}\hfil 
\begin{subfigure}{0.25\textwidth}
  \includegraphics[width=\linewidth]{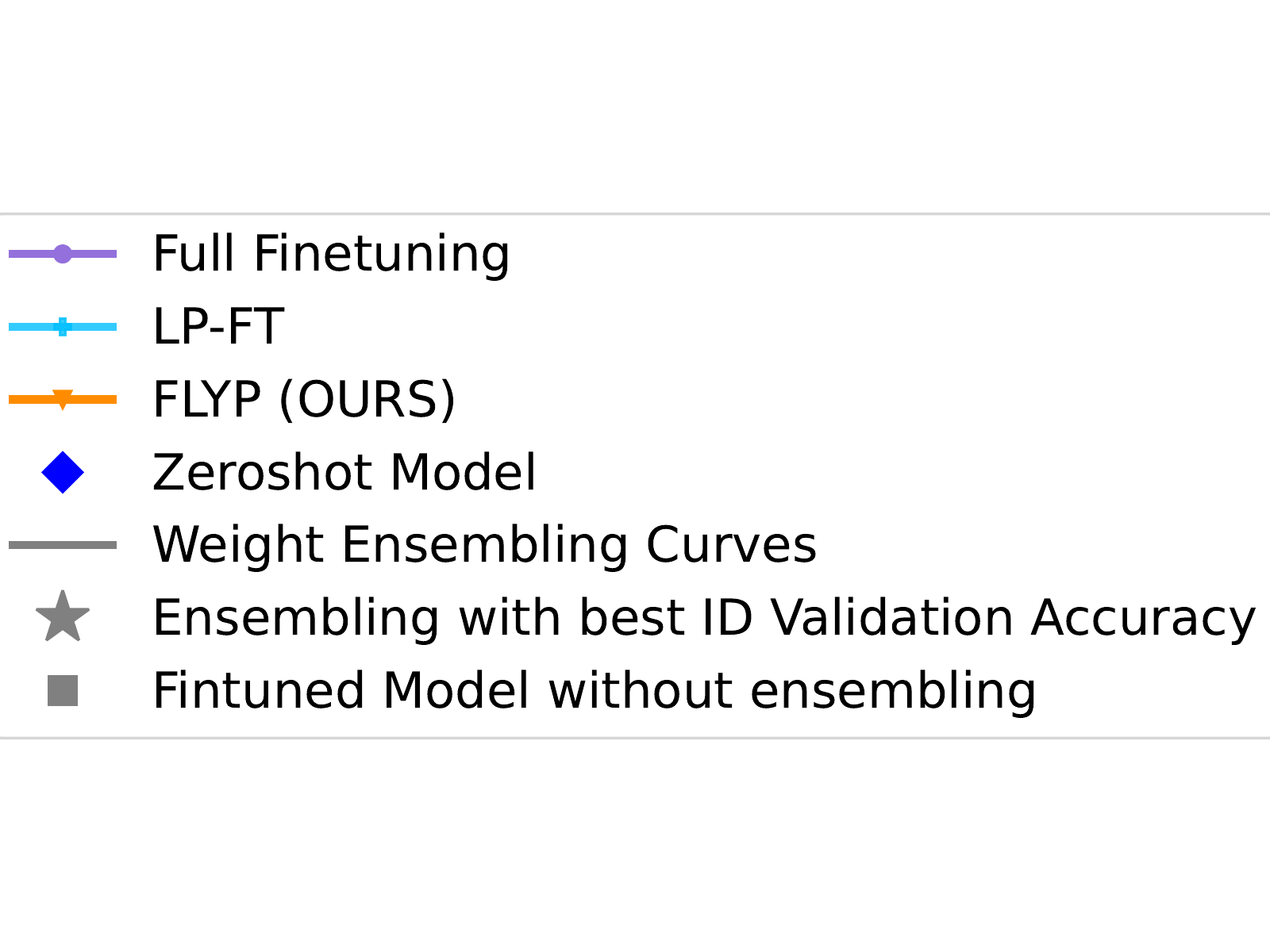}
  \caption{Legend}
  \label{fig:fewshot_imnet_32}
\end{subfigure}
\caption{We evaluate \method on few-shot classification on ImageNet, where it outperforms all the baselines with weight ensembling, giving gains of $1.5\%$ in $4$-shot classification and $0.8\%$ in $16$-shot classification over LP-FT.}
\label{fig:fewshot_imagenet}
\vspace{-1em}
\end{figure*}

\vspace{-1em}
\subsubsection{Few-shot classification on Imagenet }
\vspace{-1pt}
Figure~\ref{fig:fewshot_imagenet} plots the average OOD accuracy on $5$ Imagenet-shifts datasets versus the ImageNet ID accuracy, under $4, 16$ and $32$ shot classification. \method outperforms the baselines under all the $3$ settings. For example, \method has $1.5\%$ higher OOD accuracy compared to LP-FT in $4$-shot classification and $0.8\%$ higher OOD accuracy under $16$-shot classification, with weight ensembling.
\subsection{Transfer Learning}
\label{sec:eval_transfer}
\method generalizes well across various transfer learning datasets, giving \emph{consistently} strong empirical performance. Transfer learning is a more general setup where we are given a downstream labeled dataset to finetune and the goal is to achieve a high in-distribution performance. Table~\ref{tab:transfer_learning} compares \method with baselines on $6$ transfer datasets: PatchCamelyon, CalTech101, StanfordCars, Flowers102, ImageNet and iWILDCam. Our proposed approach \method generalizes well on all the $6$ datasets. For example, \method has $2.5\%$ higher accuracy than the most competitive baseline on iWILDCam and $1.2\%$ higher on CalTech101.

\begin{table}[]
\renewcommand{\arraystretch}{1}
\centering
\begin{minipage}[t!]{0.75\textwidth}%
\resizebox{1.0\textwidth}{!}{
\begin{tabular}{@{}ccccccc@{}}
\toprule
Methods  & PCAM                & CalTech             & Cars                & Flowers             & ImageNet            & iWILD               \\ \midrule
Zeroshot & 56.49 (-)           & 87.7 (-)            & 64.4 (-)            & 71.2 (-)            & 68.3 (-)            & 8.7 (-)             \\
LP       & 82.6 (0.1)          & 94.8 (0.0)          & 83.1 (0.0)          & 95.9 (0.0)          & 79.9 (0.0)          & 44.5 (0.6)          \\
FT       & 89.1 (1.3)          & 97.2 (0.1)          & 84.4 (0.3)          & 90.4 (0.5)          & 81.4 (0.1)          & 48.1 (0.5)          \\
LP-FT    & 89.0 (0.6)          & 96.9 (0.6)          & 89.4 (0.1)          & \textbf{97.9 (0.1)} & 81.8 (0.1)          & 49.7 (0.5)          \\ \midrule
\rowcolor[HTML]{C0C0C0} 
FLYP     & \textbf{90.3 (0.3)} & \textbf{97.6 (0.1)} & \textbf{89.6 (0.3)} & 97.7 (0.1)          & \textbf{82.6 (0.0)} & \textbf{52.2 (0.6)} \\ \bottomrule
\end{tabular}
}
\end{minipage}
\caption{We evaluate our proposed approach \method on 6 transfer learning datasets. \method gives \emph{consistently} strong empirical performance, outperforming the baselines on $5/6$ datasets considered.}
\label{tab:transfer_learning}
\vspace{-1em}
\end{table}
\section{Ablations: Why does FLYP improve performance?}\label{sec:discussion}
In this section, we attempt to shed light on what makes FLYP work so well? Our hypothesis is that this is because FLYP's finetuning objective \emph{exactly} matches the pretraining objective. We attempt to test this hypothesis and rule out a few alternate candidate hypotheses below.

First, we note that the contrastive loss we use allows for the possibility of \emph{class collisions}: multiple examples in the minibatch with the same class, yet where the contrastive loss only encourages similarity between each sample and its \emph{one} correspond sampled caption. However, we observe that correcting the class collisions in FLYP's finetuning process (which apriori seems like it should improve performance), does not really give any gains and actually slightly hurts the performance. This suggests that it is indeed important that the finetuning process matches the pretraining process, even when this comes at the detriment of other desiderata.  

Next, when compared to standard finetuning methods, FLYP has three important changes: (i) FLYP updates both the image and language encoders while incorporating structure from the text embeddings of the corresponding labels (e.g., some classes are closer than others), (ii) FLYP uses the contrastive loss rather than the cross-entropy loss, and (iii) FLYP samples prompts from a distribution which adds additional stochasticity in the finetuning process compared to other approaches. We perform experiments to tease out the role of each of these components below and see if any of them in isolation can account for all the gains.

\paragraph{Removing class collisions in FLYP's contrastive loss.} \method uses contrastive loss to finetune CLIP, which pushes representations of every image away from those of the text-descriptions of other examples in mini-batch. However, a mini-batch can potentially have multiple samples from the same class (especially so when we have a small number of classes)---the loss thus has terms that contrast an image from the text description corresponding to the correct class, which seems wasteful. 

First, we note that despite collisions, \method outperforms baselines on both the binary classification datasets (PatchCamelyon and SST2) we considered, as shown in Table~\ref{tab:fewshot_transfer} and Table~\ref{tab:transfer_learning}. We further experimented on a 10-class classification dataset of euroSAT~\citep{eurosat}, where \method again gives an accuracy of $99\%$ on-par with the baselines.

Furthermore, we attempted to resolve this ``collision'' by simply masking out such terms in the loss. However, we observed that on the transfer learning task of PatchCamelyon (with 2 classes), \method + masking performs $1.3\%$ \emph{worse} than naive FLYP which does not correct for class collisions. This shows that it is important to exactly match the pretraining process and subtle changes that seem like they should improve performance don't really give any gains. 

\paragraph{Replacing contrastive loss of FLYP with cross-entropy.} Vanilla full finetuning (FFT) does not update the language encoder during finetuning. 
Consequently, vanilla FFT does not incorporate structure from the text-embeddings of the corresponding classnames (e.g. some classes might be more closer to each other). One might wonder if FLYP's gains are entirely due to updating the language encoder? 
Consider a version of our approach \ours-CE, where we simply replace the \ours's objective with cross-entropy loss. Specifically, the embedding of text-descriptions for all the classes is used as the linear head, projecting image embeddings to the class predictions. We then use cross-entropy loss to finetune both the encoders. On Imagenet, as shown in Figure~\ref{fig:ablation} (pink curve), \ours-CE performs much worse than \method, which uses the contrastive loss. On iWILDCam, \ours-CE is $2\%$ worse ID and $0.6\%$ worse OOD than \ours. This highlights that the choice of loss function used matters indeed, and the empirical gains of our approach are not only because it updates the language encoder or uses the structure in text-descriptions of classnames.

\paragraph{Updating image embeddings via contrastive loss.} The above finding raises the corresponding alterantive question: are the gains of FLYP simply because of the contrastive loss \emph{alone}? I.e., would they happen even \emph{without} updating the language component of the model, but just updating the vision component.  However, we find that it is important to also update the language encoder. Keeping the language encoder frozen when using \method, deteriorates the performance on both ImageNet (Figure~\ref{fig:ablation}, dark blue curve) and iWILDCam (drop of $2.5\%$ ID and $1.3\%$ OOD).

\paragraph{Number of prompt templates.} Finally, we test whether FLYP's gains come from sampling different text prompts while training. To do so, we experiment on ImageNet using a single text-description template instead of 80 templates as used in ~\citet{wiseft}, and observe that finetuning accuracy of \method is not affected by the number of text-templates used. We also note that for datasets like PatchCamelyon, SST2 and Flowers, experiments in Section~\ref{sec:experiment} use only a single template as provided in ~\citet{clip}, and still outperform all the baselines.

\begin{figure}[t]
    \centering 
\begin{subfigure}{0.4\textwidth}
  \includegraphics[width=\linewidth]{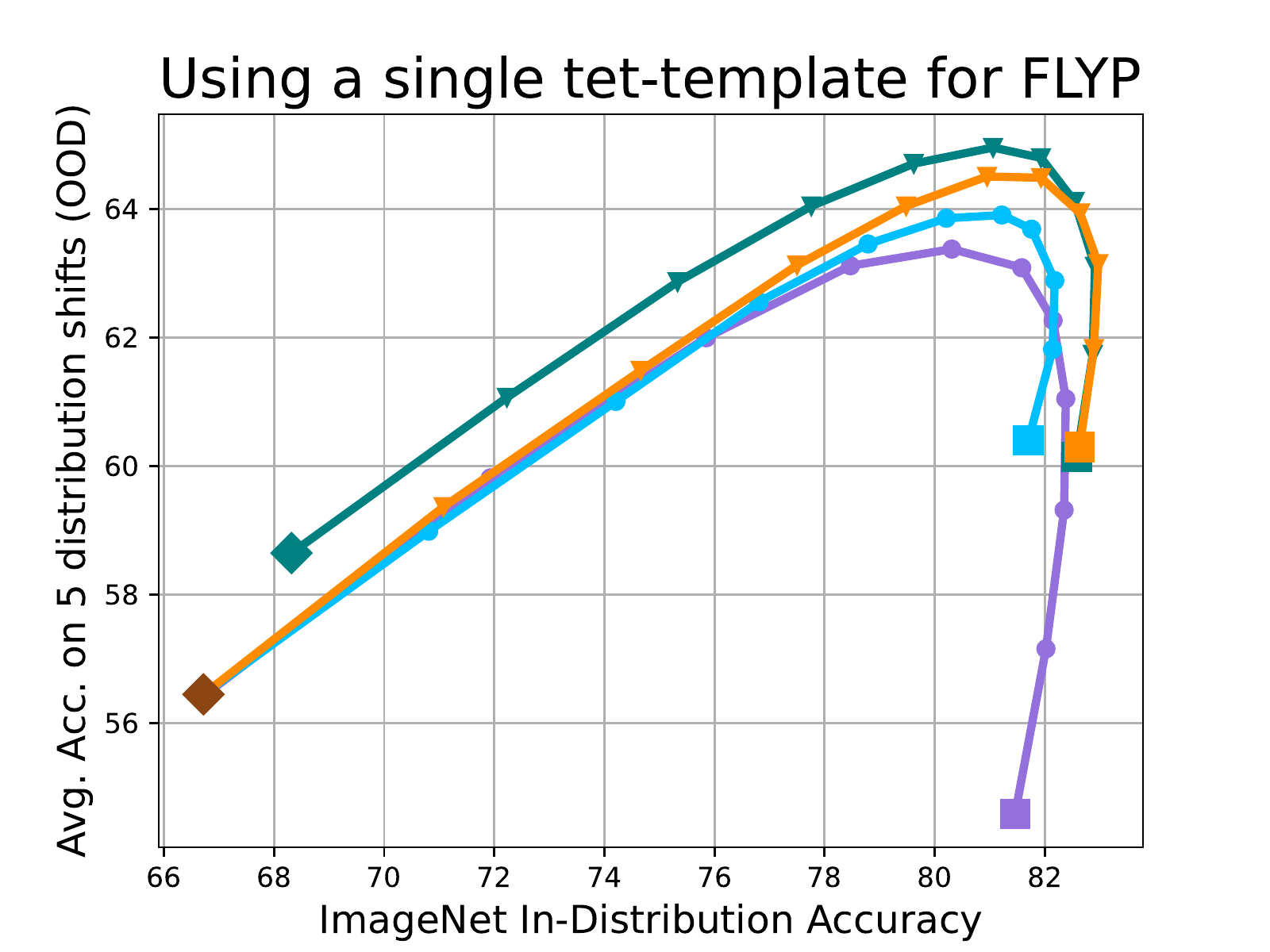}
  \label{fig:imagenet_single_temp_fig}
\end{subfigure}\hfil 
\begin{subfigure}{0.4\textwidth}
  \includegraphics[width=\linewidth]{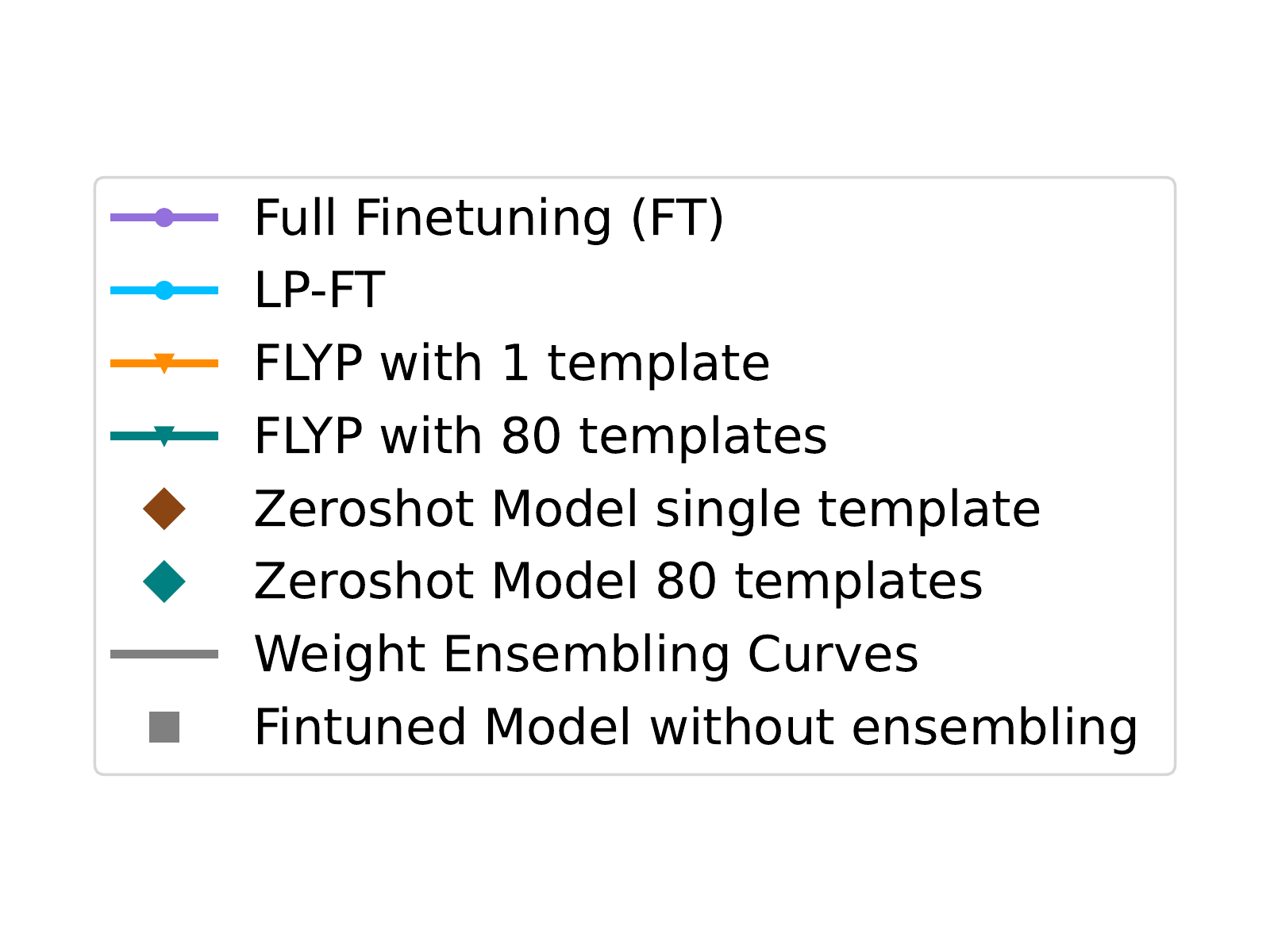}
  \label{fig:imagenet_single_temp_legend}
\end{subfigure}\hfil 
\caption{\method's performance is un-affected by the number of text-templates used during finetuning process. Here we compare using a single template versus 80 templates for text-descriptions on ImageNet dataset. Observe that \method with single template gives the same ID and OOD accuracy as \method with 80 templates, without ensembling. Note that the zeroshot model is also constructed using a single template, which causes a drop in it's accuracy, similar to the observations in ~\citet{clip}.}
\label{fig:imagenet_single_temp}
\vspace{-1em}
\end{figure}

Recall from Section~\ref{sec:method} that for every given image-label pair $(x,y)$, we construct a corresponding image-text pair $(I,T)$, where $T$ is sampled from a set of text-descriptions $\sT_y$. For example, for every class, possible text descriptions in $\sT_y$ can be ``a photo of a \{class\}'', ``a painting of a \{class\}'', ``\{class\} in wild'', etc. For all our experiments on all the datasets, we use the same text-templates as used in CLIP~\citep{clip} and WiseFT~\citep{wiseft}. Figure~\ref{fig:imagenet_single_temp} compares \method with baselines when a only a single text-description
template is used on ImageNet (the default template of ``a photo of a \{class\}'' as given in ~\citet{clip}). We observe no change in the accuracy of \method both ID and OOD (without zeroshot ensembling) when using a single template or 80 templates. Note that since the corresponding zeroshot head is also constructed using a single text-description, their is a slight drop in ensembling accuracy (due to a decrease in the zeroshot model's performance) for all the baselines as expected.

To summarize, FLYP's gains seem to come from exactly matching the pretraining process---no individual change accounts for all the gains.  In Appendix A, we talk about some more variations of \method, where we show that adding cross-entropy loss to \ours's objective (joint optimization of both) degrades the performance (with weight ensembling). Batch-size has been observed to cause variations in performance when using contrastive loss ~\citep{chen2020simclr}. All the experiments in this work used a fixed batch-size of 512 for ImageNet and 256 for rest of the datasets. However, in Appendix A, we consider the variations in performance when using a lower batch-size, observing that smaller batch size gives similar ID accuracy as larger batch-size, but a slightly lower OOD accuracy on ImageNet and FMOW and similar OOD accuracy on iWILDCam.

\section{Related works}\label{sec:related_works}
\paragraph{Standard finetuning methods.} The most standard approaches for finetuning pretrained models are linear probing and full finetuning (Section~\ref{sec:preliminary}). They have been used for supervised pretrained models~\citep{kornblith2019better, zhai2020largescale,Kolesnikov2020BigT}, self-supervised vision models such as SimCLR~\citep{chen2020simclr}, MoCo~\citep{chen2020improved,chen2021empirical}, and vision-text models such as CLIP~\citep{radford2021clip,wortsman2021robust}. For vision-text models, we find that \method outperforms these standard approaches across a variety of settings and datasets while being just as simple (or even simpler) to implement.
\vspace{-1em}
\paragraph{Recent innovations in robust finetuning of vision models.} Image-text pretrained models such as CLIP offer large improvements in robustness (OOD accuracy), and so downstream robustness has been a focus of the original CLIP paper~\citep{radford2021clip} and follow-up works~\citep{andreassen2021evolution,wortsman2021robust,kumar2022calibrated} and also in  \citep{kumar2022finetuning,zhang2022contrastive,zhou2022conditional}. It has been widely observed that standard finetuning approaches deteriorate robustness over zero-shot and a number of improvements have been proposed. The ``state-of-the-art'' approach from the literature is a combination of two recent works:~\citep{lpft} (LP-FT) and~\citep{wiseft} (weight ensembling). 

Ensembling finetuned and zeroshot models ( weight averaging) has been shown to improve both accuracy and robustness~\citep{wiseft, kumar2022calibrated, wortsman2022modelsoups}. In particular,~\citet{wortsman2022modelsoups} show that ensembling several ($>70$) models of ViT-G architecture with LP-FT leads to state-of-the-art (highest reported number) accuracies on ImageNet, WILDS-iWildCam, and WILDS-FMoW. In our work, for computational reasons, we compare with ensembling two models on ViT-B/16 and find that FLYP consistently outperforms weight averaging with LP-FT. \looseness=-1

LP-FT is a two-stage process of linear probing followed by finetuning, that outperforms other proposed alternatives~\citep{kumar2022finetuning} involving explicit regularization to pretrained weights (such as in~\citep{l2sp, xuhong2018explicit}) or selectively updating only a few parameters~\citep{zhang2020sidetuning, guo2019spottune}.
Other fine-tuning approaches use computationally intensive smoothness regularizers~\citep{zhu2020freelb, jiang2021smart} or keep around relevant pretraining data~\citep{ge2017borrowing}. 
Several other finetuning alternatives focus on improving efficiency~\citep{gao2021clip,zhang2021tip,zhou2022conditional,learn_prompt}.
\vspace{-1em}
\paragraph{Supervised learning via contrastive loss.} We advocate finetuning zeroshot models like they were pretrained---via a contrastive loss. This general idea of incorporating contrastive loss during supervised learning has been explored in other related but \emph{different} contexts as a way to regularize supervised learning.~\citet{khosla2020supervised} studies the standard \emph{fully} supervised setting (without a pretrained model),~\citep{gunel2020supervised} studies the NLP setting of finetuning large language models,~\citep{zhang2021unleashing} studies vision only models. Apart from the different settings and focuses, an important difference in methodology is that these works use the contrastive loss as an additional regularizer \emph{in addition} to cross-entropy. However, in our experiments, we observe that adding cross-entropy to \method loss degrades performance. Our work also compares the two loss functions when used independently, and shows that using the contrastive loss achieves higher accuracies than the cross-entropy loss.
\vspace{-1em}
\looseness=-1
\paragraph{Finetuning to match pretraining.} To the best of our knowledge, we are the first to document and point out that simply matching the pretraining loss while finetuning outperforms more complex alternatives across several settings and datasets. However, this idea of matching the pretraining and finetuning process has been used in T5 models~\citep{raffel2019exploring} where all NLP tasks (including pretraining objectives) are converted to a text-text task. Here, the goal was to obtain a unified framework to compare different methods. Finally, we note that~\citet{pham2021combined} seem to use a similar finetuning approach, but they do not compare to or report gains over other finetuning approaches, and they use a non-open-source model, so we cannot compare.

\vspace{-5pt}
\section{Conclusion}

In this work, we have advocated for the Finetune Like You Pretrain (FLYP) method for finetuing zero-shot vision classifiers.  The basic approach is extremely straightfowrad: when finetuning a prompt-based classifier from labeled data, we use the same contrastive loss used for pre-training, rather than the typical cross-entropy loss.  Our main contribution in this paper is to show that despite its simplicity, FLYP consistently outperforms alternative approaches that have been developed specifically for such finetuning.  Thus, we believe this paper offers strong evidence that this approach should become a ``standard'' baseline for the evaluation of finetuning image-text models. Going forward, it would be valuable to evaluate this principle of matching the finetuning loss to pretraining in other zero and few-shot learning settings, particularly in combination with other proposed heuristics (such as selectively updating parameters). Understanding this curious phenomenon where a finetuning procedure that just naively matches the pretraining loss outperforms more complex counterparts remains an open challenge.



\clearpage
{\small
\bibliographystyle{plainnat}
\bibliography{ref,all}
}

\clearpage
\appendix
\section{Additional experiments}\label{sec:appendix_a}

\subsection{Adding cross-entropy loss to \method}
\begin{figure}[t]
    \centering 
\begin{subfigure}{0.4\textwidth}
  \includegraphics[width=\linewidth]{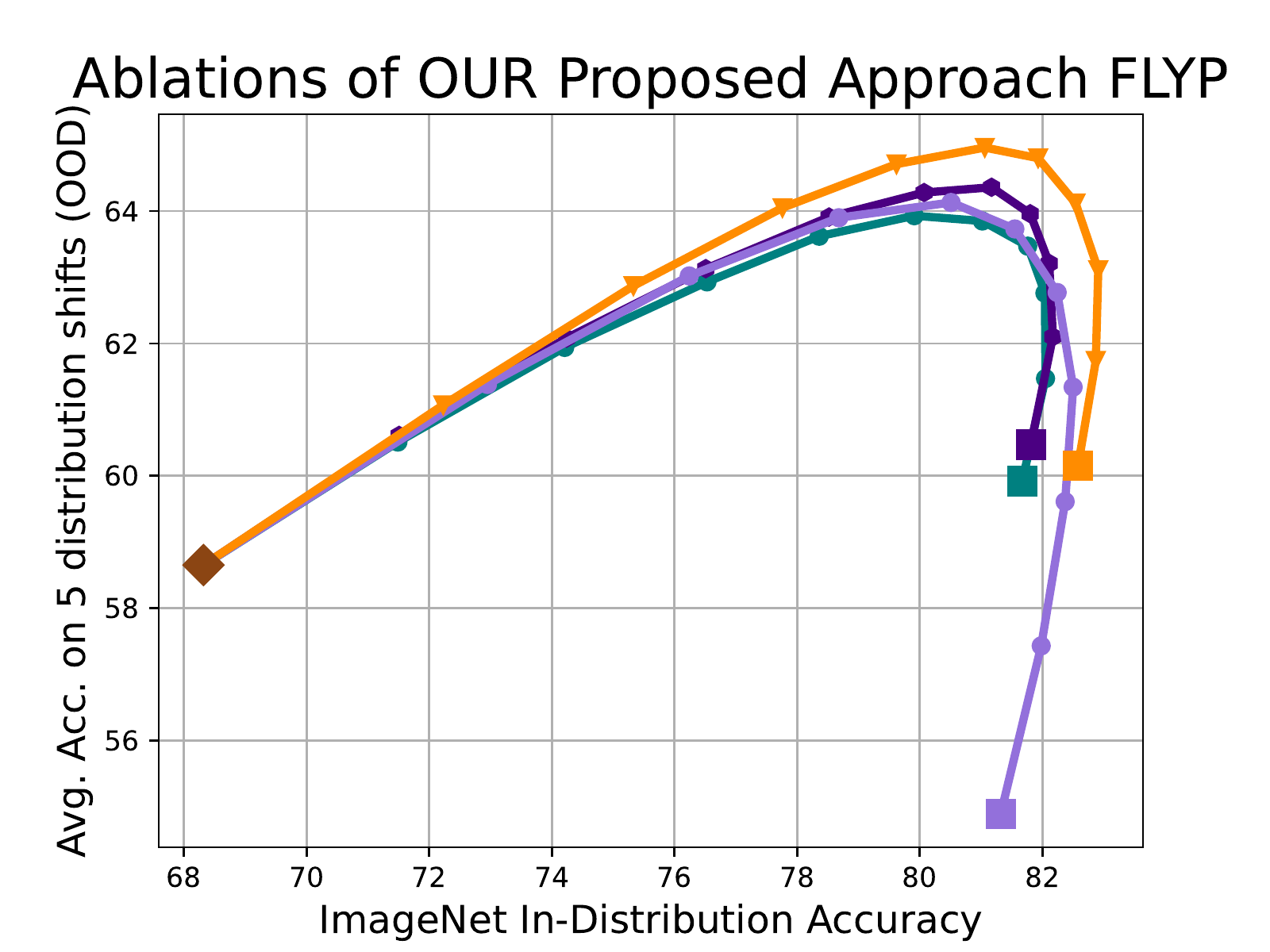}
  \caption{\centering Adding cross-entropy to \method}
  \label{fig:ce_plus_flyp}
\end{subfigure}\hfil 
\begin{subfigure}{0.4\textwidth}
  \includegraphics[width=\linewidth]{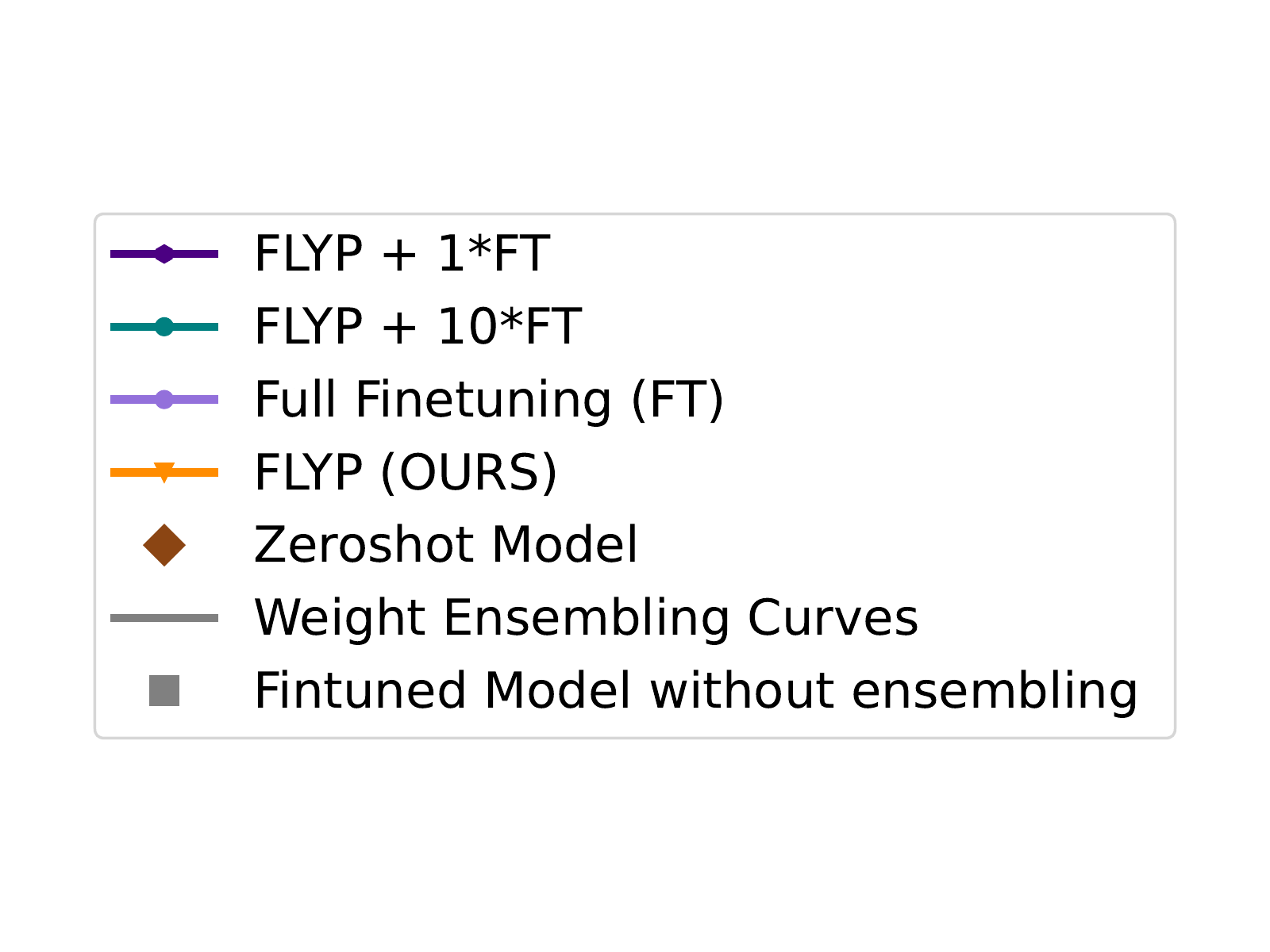}
  \caption{Legend}
  \label{fig:legend_ce_plus_flyp}
\end{subfigure}\hfil 
\caption{\centering Adding cross-entropy loss to \ours's objective degrades the performance on ImageNet.}
\label{fig:ablation2}
\vspace{-1em}
\end{figure}

In Section~\ref{sec:discussion} we observed that updating both the encoders using cross-entropy loss degrades the performance. Here we compare the performance of \method when cross-entropy loss is added to \ours's objective (i.e. the contrastive loss). On ImageNet, as shown in Figure~\ref{fig:ablation2}, the weight ensembling curve for \method (orange) completely dominates (lies above and to the right) those of when cross-entropy loss is added to \method's objective under various regularization strengths. Similarly, on iWILDCam, adding cross-entropy loss (in equal weightage) leads to a \emph{drop of} $2.64\%$ ID and $0.5\%$ OOD, as shown in Table~\ref{tab:iwildcam_ablation2}. The performance degrades further, as the weight of cross-entropy loss is increased.

\begin{table}[t]
\renewcommand{\arraystretch}{1}
\centering
\begin{minipage}[b]{0.4\textwidth}%
\centering
\resizebox{1.0\textwidth}{!}{
\begin{tabular}{@{}ccccc@{}}
\toprule
FLYP with      & \multicolumn{2}{c}{iWILDCam} & \multicolumn{2}{c}{FMOW} \\ \cmidrule(l){2-5} 
CE loss        & ID            & OOD          & ID          & OOD        \\ \midrule
\rowcolor[HTML]{C0C0C0} 
FLYP           & 53.03         & 38.09        & 68.56       & 40.76      \\ 
FLYP + FT      & 50.39         & 37.59        & 69.02       & 39.45      \\
FLYP + 10 * FT & 48            & 35.35        & 68.45       & 39.45      \\
FT             & 47.99         & 34.77        & 68.47       & 39.53      \\ \bottomrule
\end{tabular}
}
\end{minipage}
\caption{Adding cross-entropy loss to \ours's objective (with various regularization) decreases the performance both ID and OOD.}
\label{tab:iwildcam_ablation2}
\vspace{-1em}
\end{table}

\subsection{Effect of batch-size on \method}
Batch-size has been observed to cause variations in performance when using contrastive loss ~\citep{chen2020simclr}. As mentioned in Section~\ref{sec:experiment}, we use a fixed batch size of 512 for ImageNet and 256 for rest of the datasets. We perform additional experiments with a lower batch-size (half of previous). On ImageNet, a lower batch-size of 256 gives similar ID accuracy as batch-size of 512, however we observe a slight drop of $1\%$ in the OOD accuracy. On iWILDCam, with a lower batch-size of 128, we get similar ID and OOD accuracy as the batch-size of 256. However, on FMOW, we again observe a drop in OOD accuracy of $0.5\%$.

\subsection{Few-shot classification using CLIP ViT-L/14}
In Section~\ref{sec:binary_few_shot}, we considered a challenging task of binary few-shot classification on 2 datasets of PatchCamelyon and SST2, using CLIP ViT-B/16. Here we perform a similar comparison, although using a much bigger model of CLIP ViT-L/14. Table~\ref{tab:binary_few_shot_vitl} compares \method with baselines on 2 datasets of SST2 and PatchCamelyon. We observe that similar to the case of using smaller CLIP ViT-B/16 (Table~\ref{tab:fewshot_transfer}), \method outperforms the baselines when a larger model of CLIP ViT-L/14 is used. For example, under 32-shot classification, \method outperforms LP-FT by $4.2\%$ on SST2 and $3.1\%$ on PatchCamelyon.

\begin{table}[t]
\renewcommand{\arraystretch}{1}
\centering
\begin{minipage}[b]{0.75\textwidth}%
\resizebox{1.0\textwidth}{!}{
\begin{tabular}{@{}lcccccc@{}}
\toprule
\multicolumn{1}{c}{}                             & \multicolumn{3}{c}{SST2}                                        & \multicolumn{3}{c}{PatchCamelyon}                               \\ \cmidrule(l){2-7} 
\multicolumn{1}{c}{\multirow{-2}{*}{Methods}}    & 4 Shot              & 16 Shot             & 32 Shot             & 4 Shot              & 16 Shot             & 32 Shot             \\ \midrule
Zeroshot                                         & 68.9 (-)            & 68.9 (-)            & 68.9 (-)            & 62 (-)              & 62 (-)              & 62 (-)              \\
LP                                               & 69.2 (0.2)          & 70.2 (0.4)          & 71.0 (0.4)          & 66.9 (0.8)          & 71.2 (1.1)          & 74.1 (0.9)          \\
FT                                               & 69.3 (0.1)          & 69.5 (0.2)          & 70.0 (0.4)          & 65.9 (0.7)          & 69.9 (0.7)          & 71.8 (0.8)          \\
LP-FT                                            & 69.3 (0.3)          & 70.7 (0.4)          & 71.3 (0.4)          & 67.5 (0.8)          & 72.9 (1.0)          & 76.0 (0.5)          \\ \midrule
\rowcolor[HTML]{C0C0C0} 
\multicolumn{1}{c}{\cellcolor[HTML]{C0C0C0}FLYP} & \textbf{69.8 (0.7)} & \textbf{73.2 (0.8)} & \textbf{75.5 (0.6)} & \textbf{67.8 (1.2)} & \textbf{75.8 (0.9)} & \textbf{79.1 (0.7)} \\ \bottomrule
\end{tabular}
}
\end{minipage}
\caption{ Binary few-shot classification using CLIP ViT-L/14. \method continues to outperforms the baselines. For example, under 32-shot classification, \method outperforms LP-FT by $4.2\%$ on SST2 and $3.1\%$ on PatchCamelyon.}
\label{tab:binary_few_shot_vitl}
\vspace{-1em}
\end{table}

\subsection{ImageNet Distribution Shifts - Detailed Results}
Table~\ref{tab:imagenet_detailed_shifts} gives the detailed results on each individual associated distribution shifts with the ImageNet dataset, in the same experiment setting as Section~\ref{sec:eval_dist_shift}. Observe that with weight ensembling, \method consistently outperforms the baselines across all the distribution shifts.

\begin{table*}[t!]
\centering
\resizebox{0.95\textwidth}{!}{
\begin{tabular}{@{}ccccccccccccccc@{}}
\toprule
                          &                                                            & \multicolumn{5}{c}{ImageNet Distribution Shifts}                                                                           &                                                   &                                                            & \multicolumn{5}{c}{ImageNet Distribution Shifts}                                                                           &               \\ \cmidrule(l){2-15} 
\multirow{-2}{*}{Methods} & ID                                                         & Im-V2         & Im-R          & Im-A          & Sketch        & \cellcolor[HTML]{FFFFFF}ObjectNet                          & \multicolumn{1}{c|}{Avg. OOD}                     & ID                                                         & Im-V2         & Im-R          & Im-A          & Sketch        & \cellcolor[HTML]{FFFFFF}ObjectNet                          & Avg. OOD      \\ \midrule
Zeroshot                  & \multicolumn{1}{c|}{68.3}                                  & 61.9          & 77.7          & 50.0          & 48.3          & \multicolumn{1}{c|}{55.4}                                  & \multicolumn{1}{c|}{58.7}                         & \multicolumn{1}{c|}{68.3}                                  & 61.9          & 77.7          & 50.0          & 48.3          & \multicolumn{1}{c|}{55.4}                                  & 58.7          \\
LP                        & \multicolumn{1}{c|}{79.9}                                  & 69.8          & 70.8          & 46.4          & 46.9          & \multicolumn{1}{c|}{52.1}                                  & \multicolumn{1}{c|}{57.2}                         & \multicolumn{1}{c|}{80.0}                                  & 70.3          & 72.4          & 47.8          & 48.1          & \multicolumn{1}{c|}{52.8}                                  & 58.3          \\
FT                        & \multicolumn{1}{c|}{81.3}                                  & 71.2          & 66.1          & 37.8          & 46.1          & \multicolumn{1}{c|}{53.3}                                  & \multicolumn{1}{c|}{54.9}                         & \multicolumn{1}{c|}{82.5}                                  & 72.8          & 74.9          & 48.1          & 51.9          & \multicolumn{1}{c|}{59.0}                                  & 61.3          \\
L2-SP                     & \multicolumn{1}{c|}{81.7}                                  & 71.8          & 70.0          & 42.5          & 48.5          & \multicolumn{1}{c|}{56.2}                                  & \multicolumn{1}{c|}{57.8}                         & \multicolumn{1}{c|}{82.2}                                  & 72.9          & 75.1          & 48.6          & 51.4          & \multicolumn{1}{c|}{58.9}                                  & 61.4          \\
LP-FT                     & \multicolumn{1}{c|}{81.7}                                  & 72.1          & \textbf{73.5} & 47.6          & \textbf{50.3} & \multicolumn{1}{c|}{58.2}                                  & \multicolumn{1}{c|}{\textbf{60.3}}                & \multicolumn{1}{c|}{82.1}                                  & 72.8          & 75.3          & 50.1          & 51.7          & \multicolumn{1}{c|}{59.2}                                  & 61.8          \\ \midrule
\rowcolor[HTML]{C0C0C0} 
FLYP                      & \multicolumn{1}{c|}{\cellcolor[HTML]{C0C0C0}\textbf{82.6}} & \textbf{73.0} & 71.4          & \textbf{48.1} & 49.6          & \multicolumn{1}{c|}{\cellcolor[HTML]{C0C0C0}\textbf{58.7}} & \multicolumn{1}{c|}{\cellcolor[HTML]{C0C0C0}60.2} & \multicolumn{1}{c|}{\cellcolor[HTML]{C0C0C0}\textbf{82.9}} & \textbf{73.5} & \textbf{76.0} & \textbf{53.0} & \textbf{52.3} & \multicolumn{1}{c|}{\cellcolor[HTML]{C0C0C0}\textbf{60.8}} & \textbf{63.1} \\ \bottomrule
\end{tabular}
}
\caption{\method compared with baselines on each of the associated distribution shifts with the ImageNet dataset. Note that with weight ensembling, \method consistently outperforms all the baselines on all the distribution shift benchmarks as well as ID.}
\label{tab:imagenet_detailed_shifts}
\vspace{-1em}
\end{table*}

\section{Experimental details}\label{sec:appendix_b}
\label{app:experiment_details}

\subsection{Hyper-parameter Sweep Details}
 For all algorithms on all the datasets (apart from ImageNet), we perform a hyper-parameter sweep over 5 learning rates in $\{1e^{-2}, 1e^{-3}, ..., 1e^{-6}\}$ and 5 weight decay in $\{$0.0, 0.1, ..., 0.4$\}$, with a fixed batch-size of $256$. For ImageNet, due to computational cost, we perform a sweep over 3 learning rates in $\{$1e-4, 1e-5, 1e-6$\}$ and 2 weight decay in $\{$0.0, 0.1$\}$ and use a batch-size of $512$. L2-SP requires tuning a additional regularization term weight $\lambda \in \{1e^{-1}, 1e^{-2}, ..., 1e^{-4}\}$.

 We early stop and choose the best hyper-parameter based on the ID Validation accuracy. For the datasets which do not have a publicly available validation split, we split the training dataset in 80:20 ratio to make a training and a validation set. In all the cases, note that the OOD dataset is used only for evaluation purposes.

For the $k$ few-shot classification setting (where $k \in \{4,16,32\}$), we randomly sample $k$ training and $k$ validation points from the respective full datasets and repeat the process 50 times, due to increased variance caused by a small training and validation set. We finally report the mean test accuracy over the $50$ runs, corresponding to the hyper-parameter with the lowest mean validation loss over the $50$ runs.

\end{document}